\documentclass[10pt,twocolumn,letterpaper]{article}

\usepackage{wacv}
\usepackage{pdfpages}
\usepackage{times}
\usepackage{epsfig}
\usepackage{graphicx}
\usepackage{amsmath}
\usepackage{amssymb}
\usepackage[ruled,lined]{algorithm2e}

\wacvfinalcopy 

\def\wacvPaperID{39} 

\ifwacvfinal\fi
\begin{document}

\renewcommand{\dblfloatpagefraction}{1}
\renewcommand{\textfraction}{0}
\renewcommand{\dblfloatsep}{5pt}
\renewcommand{\dbltextfloatsep}{1ex}
\renewcommand{\dbltopfraction}{1}
\renewcommand{\intextsep}{5pt}
\renewcommand{\floatpagefraction}{1}
\renewcommand{\floatsep}{5pt}
\renewcommand{\textfloatsep}{1ex}
\renewcommand{\topfraction}{1}
\renewcommand{\abovecaptionskip}{2pt}
\let\OldCaption=\caption
\renewcommand{\caption}[1]{\small\OldCaption{\em#1}}

\makeatletter
\def\@normalsize{\@setsize\normalsize{10pt}\xpt\@xpt
\abovedisplayskip 10pt plus2pt minus5pt\belowdisplayskip
\abovedisplayskip \abovedisplayshortskip \z@
plus3pt\belowdisplayshortskip 6pt plus3pt
minus3pt\let\@listi\@listI}
\def\subsize{\@setsize\subsize{12pt}\xipt\@xipt}
\def\section{\@startsection {section}{1}{\z@}{1.0ex plus
1ex minus .2ex}{.2ex plus .2ex}{\large\bf}}
\def\subsection{\@startsection {subsection}{2}{\z@}{.2ex
plus 1ex} {.2ex plus .2ex}{\subsize\bf}} \makeatother

\newcommand{\Section}[1]{\section{\hskip -1em.~~#1}}
\newcommand{\SubSection}[1]{\subsection{\hskip -1em.~~#1}}
\def\@listI{%
 \leftmargin\leftmargini
 \partopsep 0pt
 \parsep 0pt
 \topsep 0pt
 \itemsep pt
 \relax
} \long\def\@makecaption#1#2{
 \vskip -5pt
 \setbox\@tempboxa\hbox{\small{#1\,:\,#2}}
  \ifdim \wd\@tempboxa >\hsize \unhbox\@tempboxa\par \else
  \hbox to\hsize{\hfil\box\@tempboxa\hfil}
\fi \vskip -0.2cm}

\jot=0pt \abovedisplayskip=3pt \belowdisplayskip=3pt
\abovedisplayshortskip=0pt \belowdisplayshortskip=0pt

\title{Unsupervised Joint Mining of Deep Features and Image Labels \\ for Large-scale Radiology Image Categorization and Scene Recognition }
\author{Xiaosong Wang, 
	Le Lu, 
	Hoo-chang Shin, 
	Lauren Kim,	
	Mohammadhadi Bagheri,\\
	Isabella Nogues, 
	Jianhua Yao, 
	Ronald M. Summers\\
	Department of Radiology and Imaging Sciences, National Institutes of Health Clinical Center,\\ 
	10 Center Drive, Bethesda, MD 20892\\
	{\tt\small \{xiaosong.wang,le.lu,lauren.kim2,mohammad.bagheri,isabella.nogues\}@nih.gov,}\\ {\tt\small jyao@cc.nih.gov, rms@nih.gov}
}
\maketitle
\ifwacvfinal\thispagestyle{empty}\fi

\begin{abstract}

The recent rapid and tremendous success of deep convolutional neural networks (CNN) on many challenging computer vision tasks largely derives from the accessibility of the well-annotated ImageNet and PASCAL VOC datasets. Nevertheless, unsupervised image categorization (i.e., without the ground-truth labeling) is much less investigated, yet critically important and difficult when annotations are extremely hard to obtain in the conventional way of ``Google Search'' and crowd sourcing. We address this problem by presenting a looped deep pseudo-task optimization (LDPO) framework for joint mining of deep CNN features and image labels. Our method is conceptually simple and rests upon the hypothesized  ``convergence'' of better labels leading to better trained CNN models which in turn feed more discriminative image representations to facilitate more meaningful clusters/labels. Our proposed method is validated in tackling two important applications: 1) Large-scale medical image annotation has always been a prohibitively expensive and easily-biased task even for well-trained radiologists. Significantly better image categorization results are achieved via our proposed approach compared to the previous state-of-the-art method. 
2) Unsupervised scene recognition on representative and publicly available datasets 
with our proposed technique is examined. The LDPO achieves excellent quantitative scene classification results. On the MIT indoor scene dataset,  it attains a clustering accuracy of $75.3$\%, compared to the state-of-the-art supervised classification accuracy of $81.0$\% 
(when both are based on the \textit{VGG-VD} model) 
.

\end{abstract}

\section{Introduction}
Deep Convolutional Neural Networks (CNNs) have demonstrated remarkable success on many challenging computer vision tasks of object recognition, detection, segmentation and scene recognition using public image datasets (e.g., Pascal VOC \cite{Everingham2015Pascal}, ImageNet ILSVRC \cite{deng2009imagenet,russakovsky2014imagenet}, MS COCO \cite{Lin2014Coco}), with significantly superior performance than previous arts, especially those non-deep methods built upon hand-crafted image features. However, the good efficacy of CNNs often comes at the cost of large amounts of annotated training data. 
ImageNet pre-trained deep CNN models \cite{jia2014caffe,krizhevsky2012imagenet,Lin2015Nin} serve an indispensable role to be bootstrapped or fine-tuned \cite{Huh2016what} for all externally-sourced data exploitation tasks \cite{Liang2015Baby,Chen2015Webly}. 

In the medical imaging domain, nevertheless, no large-scale labeled image dataset comparable to ImageNet exists (except the one in \cite{Shin2015} which is not directly comparable). Modern hospitals store vast amounts of radiological images/reports in their Picture Archiving and Communication Systems (PACS). The main challenge now lies in how to obtain or compute the ImageNet-like semantic labels given a large collection of medical images. Conventional means of collecting image labels (e.g., Google image search using the terms from WordNet ontology hierarchy \cite{Miller1995}, SUN/PLACE databases \cite{Xiao2010Sun,Zhou2014Place} or NEIL knowledge base \cite{Chen2013Neil}; followed by crowd-sourcing \cite{deng2009imagenet}) are not applicable, due to 1) unavailability of a high quality or large capacity medical image search engine, and 2) the formidable difficulties of medical annotation tasks for annotators with no clinical training. Additionally, even for well-trained radiologists, this type of ``assigning labels to images'' task is not aligned with their diagnostic routine work so that drastic inter-observer variations or inconsistency are expected. The protocols of defining image labels based on visible anatomic structures (often multiple), or pathological findings (possibly multiple) or both cues may have intrinsically high ambiguities. 

Recent semi-supervised image feature learning and self-taught image recognition techniques \cite{Singh2012DiscPat,Raina2007Self,Li2011Towards,Juneja2013Blocks,Dai2015EnProDeepFets} have advanced both supervised image classification and unsupervised clustering processes, demonstrating some promising results. Common image patches \cite{Singh2012DiscPat,LiLSH15CVPR}, object parts \cite{Juneja2013Blocks}, prototypes \cite{Dai2015EnProDeepFets,Dai2013EnPro} or spatial context \cite{doersch2015unsupervised} can be first mined amongst images of the same theme (e.g., the same indoor scene class \cite{quattoni2009recognizing}) and then concatenated to serve as the discriminative image representations for classification. All of these methods, however, require image labels in order to learn class-specific informative image representations.

In this paper, we present the \textbf{L}ooped \textbf{D}eep \textbf{P}seudo-task \textbf{O}ptimization framework (LDPO) for joint mining of image features and labels, with no prior knowledge of the image categories. The ``true'' image category labels are assumed to be latent and not directly observable. The main idea is to learn and train CNN models using pseudo-task labels (since human-annotated labels are unavailable) and iterate this process with the expectation that pseudo-task labels will gradually resemble the real image categories. This looped optimization algorithm flow starts with deep CNN feature extraction and image encoding using domain-specifically (e.g., CNN trained on radiology images and text report-derived labels \cite{Shin2015}) or generically initialized CNN models. Afterwards, the CNN-encoded image feature vectors are clustered to compute and refine image labels, then we feed the newly clustered labels to fine-tune the current CNN models. Next, the obtained more task-specific and representative deep CNN will serve as the deep image encoder in the successive iteration. This looped process will halt until a stopping criterion is met.
For medical image annotation, LDPO generated image clusters can be further interpreted by a natural language processing (NLP) based text mining system and/or a clinician \cite{Yu2015Construction}. 


Our contributions are three-fold. \textbf{1)}, 
The unsupervised joint mining of deep image features and labels via LDPO is conceptually simple and based on the hypothesized  ``convergence'' of better labels lead to better trained CNN models which in turn, offer more effective deep image features to facilitate more meaningful clustering/labels. {\it This looped property is unique to deep CNN classification-clustering models since other types of classifiers do not learn better image features simultaneously.} 
\textbf{2)}, We apply our method to the large-scale medical image auto-annotation. To the best of our knowledge, this is the first work exploiting to integrate unsupervised deep feature clustering and supervised deep label classification for self-annotating a large-scale radiology image database where the conventional means of image annotation may not be quite feasible. Our best converged model obtains the Top-1 classification accuracy of 0.8109 and Top-5 accuracy 0.9412 with 270 formed image categories. \textbf{3)}, LDPO framework is also validated through the scene recognition task where the ground-truth labels are available (only for the validation purpose). We report the 67-class clustering accuracy of $75.3$\% on the MIT-67 indoor scene dataset \cite{quattoni2009recognizing} that doubles the performance from the baseline methods (of using k-means or agglomerative clustering on the ImageNet-pretrained deep image features via AlexNet \cite{krizhevsky2012imagenet}) and is strongly close to the fully-supervised deep classification result of $81.0$\%~\cite{Cimpoi2015Deep}.

\section{Related Work} 
\label{sec-review}

{\bf Image Categorization or Auto-annotation:} Image auto-annotation task is addressed via multiple instance learning \cite{Wu2015Deep} but the target domain is restricted to a small subset (only 25 out of 1000 classes) of ImageNet \cite{deng2009imagenet} and SUN \cite{Xiao2010Sun}. \cite{Wigness2015} introduces a hierarchical set of unlabeled data clusters (spanning a spectrum of visual concept granularities) that can be efficiently labeled to produce high performance classifiers (thus less label noises than the instance-level labeling). \cite{Shin2015} first extract the sentences that depict disease referencing key images (analogous to ``key frames in videos'') via NLP from a total collection of $\sim780$K patients' radiology text reports, and 215,786 key images of 61,845 unique patients are found. Then, image categorization labels are computed using unsupervised hierarchical Bayesian document clustering, i.e., latent Dirichlet allocation (LDA) topic modeling \cite{blei2003latent}, to form 80 classes. The text-computed category information offers some coarse level of radiology semantics but appears to be limited in two aspects: 1) The classes are {\em highly unbalanced}, in which one dominating category contains 113,037 images while other classes contain a few dozens. 2) Some classes can be highly incoherent among their within-the-class image instances.  

{\bf Unsupervised and Semi-supervised Learning:} Dai \textit{et al.}~\cite{Dai2015EnProDeepFets,Dai2013EnPro} study the semi-supervised image classification and clustering on problems of texture \cite{Lazebnik2005Sparse}, small- to middle-scale object categories (e.g., Caltech-101 \cite{FeiFei2004101}) and scene recognition \cite{Quattoni2009indoor}. Ensemble projections (EP) as a rich set of visual prototypes are derived as the new image representation for clustering and recognition. Graph based approaches \cite{Liu2010Large,Kingma2014SSL} are used to link the unlabeled image instances to labeled ones (which are served as anchors) and propagate labels by the graph topology and connectiveness weights. In an unsupervised manner, Coates \textit{et al.} \cite{Coates2011} employ k-means to mine image patch filters and utilize the resulted filters for feature computation. Surrogate classes are obtained by augmenting each image patch with its geometrically transformed versions and a CNN is trained on top of these surrogate classes to generate features, as studied in \cite{Dosovitskiy2014}. \cite{yang2016joint} integrates the hierarchical agglomerative clustering process into a recurrent neural network by merging the clusters (as groups of images) iteratively toward the predefined cluster number and simultaneously updating the CNN activations for image representation. 

Our looped optimization method shares a similar concept with \cite{yang2016joint} in the joint learning of image clusters and image representations. However, it differs significantly in the following respects: 1) an unlabeled image collection can be initialized with either randomly-assigned labels or labels obtained by a pseudo-task (e.g., text topic modeling generated labels \cite{Shin2015}); 2) Our framework has the flexibility of working with any clustering function. Particularly, it employs Regularized Information Maximization (RIM \cite{Gomes2010Discriminative}) clustering to perform clustering the image (like k-means) with model selection on finding the optimal number of clusters {\em whereas only agglomerative clustering loss \cite{gowda1978agglomerative} can be integrated into the neural network model in \cite{yang2016joint}}. 3) The empirical convergence process of our LDPO method is observable and quantifiable, as described in Sec. \ref{sec:clustering}. 

{\bf Mid-level Image Representation:} Since the seminal work on discriminative image patch discovery \cite{Singh2012DiscPat}, mid-level visual elements based image representation has been explored intensively and found being effective on boosting the performance of many visual computing tasks, particularly scene recognition \cite{Singh2012DiscPat,doersch2013mid,Juneja2013Blocks,sun2013learning,li2013harvesting,bossard2014food,Dai2015EnProDeepFets,LiLSH15CVPR,Wu2015Harvesting}. A variety of mid-level visual elements can be harvested, e.g., image patches \cite{Singh2012DiscPat,doersch2013mid,LiLSH15CVPR,Wu2015Harvesting}, parts/segments \cite{Juneja2013Blocks,sun2013learning,bossard2014food}, prototypes \cite{Dai2015EnProDeepFets}, attributes \cite{shrivastava2012constrained,choi2013adding} through different learning and mining techniques, e.g., iterative optimization \cite{Singh2012DiscPat,Juneja2013Blocks}, classification and co-segmentation \cite{sun2013learning}, Multiple Instance Learning (MIL) \cite{li2013harvesting}, random forests \cite{bossard2014food}, ensemble projection\cite{Dai2015EnProDeepFets} and association rule mining \cite{LiLSH15CVPR}. Nonetheless, these methods require that images are grouped before their representations are mined inside each group, which is a form of weakly supervised learning (WSL).
 
Our work is partly related to the iterative optimization in \cite{Singh2012DiscPat,Juneja2013Blocks} that seeks to identify discriminative local visual patterns as parts and reject others, while our goal is to jointly mine better deep image representations and the labels for all images, towards iterative auto-annotation. We can integrate the association rule mining technique \cite{LiLSH15CVPR} to extract the frequent image parts (that are further used to encode image representation) into our LDPO pipeline, and report excellent unsupervised scene recognition accuracy of $75.3$\% on MIT indoor scene dataset \cite{Wu2015Harvesting,Cimpoi2015Deep,quattoni2009recognizing}. 

\section{Joint Mining of Deep Features and Labels}\label{sec-method}

Supervised or semi-supervised learning paradigms (as described in Sec. \ref{sec-review}) usually require (at least partial) image labels as a prerequisite. These lines of work, at the era of ``deep learning'', would necessitate a huge amount of data annotation efforts. For medical imaging applications, well-trained clinical professionals or physicians are in need for data labeling, instead of Amazon Mechnical Turkers in computer vision. Employing and converting the medical records stored in the PACS into image labels or tags is a highly non-trivial and unsolved NLP problem with high labeling uncertainties, observed by \cite{Shin2015Interleaved}. Our approach exploits unsupervised category discovery using empirical image cues for grouping or clustering, through an iterative optimization process of 1) deep image feature extraction and clustering; and 2) deep CNN model fine-tuning (i.e., using new labels from clustering), to update deep feature extraction in the next round. 

Without loss of generality, our method is first employed in the scenario of medical image categorization. We highlight the problem-specific settings for scene recognition task when they are different. As illustrated in Fig.~\ref{fig:flowchart:png}, the iteration begins by extracting the deep CNN image feature using either a domain-specific \cite{Shin2015} or generic ImageNet \cite{krizhevsky2012imagenet} CNN model (Sec. \ref{sec:ip}). Next, the clustering on deep feature with $k$-means or $k$-means followed by RIM is exploited (Sec. \ref{sec:clustering}). By evaluating the purity and mutual information between formed clusters in consecutive rounds, the system either terminates the current iteration (and yields converged clustering outputs); or uses the newly refined image cluster labels to train or fine-tune the CNN model in the next iteration. For medical image categorization (dashed box in Fig.~\ref{fig:flowchart:png}), LDPO-generated image clusters can be further fed into text processing. The system can extract semantically meaningful text words for each formed cluster. Furthermore, the hierarchical category relationship is built using the class confusion measures of the final converged CNN classification models (Sec. \ref{sec:hcl}). 
\begin{figure*}
	\begin{center}
		\includegraphics[width=1\linewidth]{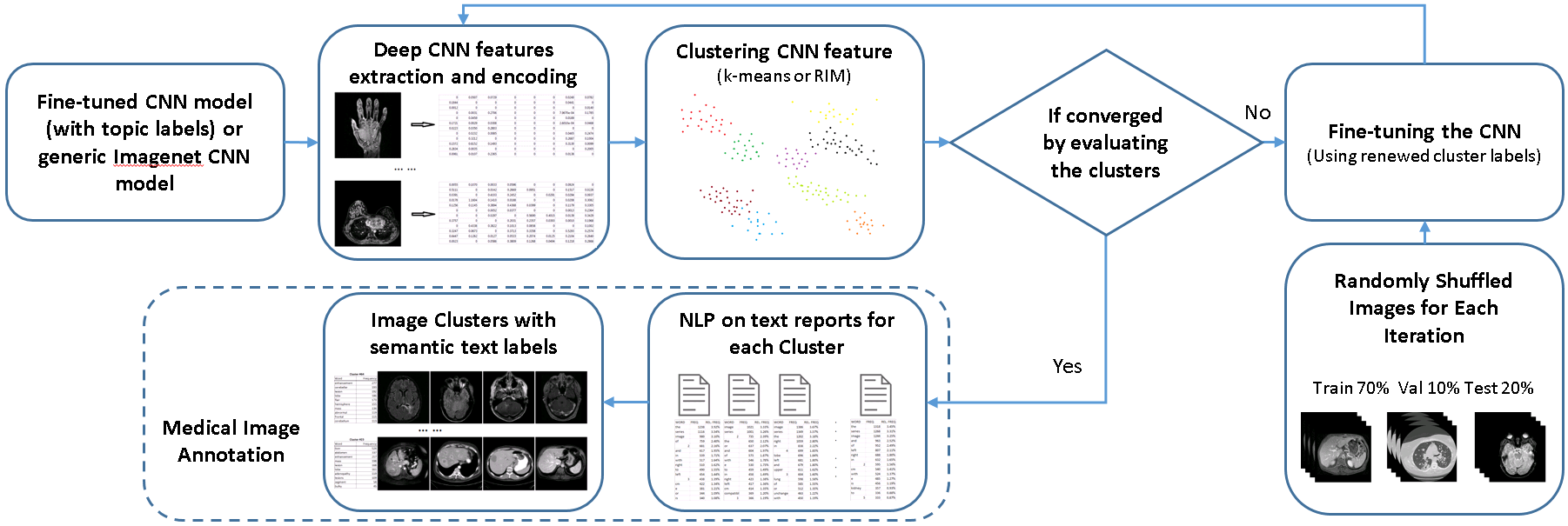}
	\end{center}
	\caption{The overview of looped deep pseudo-task optimization (LDPO) framework.}
	\label{fig:flowchart:png} 
\end{figure*}     


\subsection{Deep CNN Image Representation \& Encoding}\label{sec:ip} 

A variety of CNN models can be used in our method. We analyze the CNN activations from layers of different depths in AlexNet~\cite{krizhevsky2012imagenet}, VGGnet\cite{simonyan2014very} and GoogLeNet~\cite{Szegedy2014Going}. Pre-trained models on the ImageNet ILSVRC data are obtained from Caffe Model Zoo \cite{jia2014caffe}. We also employ the Caffe CNN implementation~\cite{jia2014caffe} to perform fine-tuning on CNNs using the key image database \cite{Shin2015,Shin2015Interleaved}.  AlexNet is a popular 7-layer CNN architecture and the extracted features from its convolutional or fully-connected layers have been broadly investigated \cite{Girshick2015RCNN,Razavian2014CNN,Karpathy2015Deep,Ng15}. 
In our experiments we harness image feature activations of the 5th convolutional layer $Conv5$ and 7th fully-connected (FC) layer $FC7$, suggested by \cite{Cimpoi2015Deep,Chatfield14}. GoogLeNet \cite{Szegedy2014Going} is a much deeper CNN architecture that comprises 9 inception modules and each module is a set of convolutional layers with multiple window sizes of $1\times1, 3\times3, 5\times5$. We utilize the deep features from the last inception layer $Inception5b$ and the final pooling layer $Pool5$. For the scene recognition task, very deep VGGNet (VGG-VD) \cite{simonyan2014very} is also employed, in addition to AlexNet. The extracted features from VGG-VD's last fully-connected layers are used for the patch-mining based image encoding. Table~\ref{tab:model} illustrates the detailed CNN layers and their activation dimensions. 

{\small\small
\begin{table}[t]
	\small
	\caption{Configurations of CNN output layers and encoding schemes (Output dimension is 4096, except Pool5 as 1024).}
	\centering
	\begin{tabular}{|l||c|c|c|}
		\hline\hline 
		{\bf CNN model}   & {\bf Layer} & {\bf Activations} & {\bf Encoding} \\
		\hline\hline
		\multicolumn{4}{|c|}{Medical Image Categorization}\\
		\hline
		{\bf AlexNet}   & FC7 & 4096 & $-$ \\ 
		\hline
		{\bf AlexNet}    & Conv5 & $(13,13,256)$ & FV+PCA \\
		\hline
		{\bf AlexNet}   & Conv5 & $(13,13,256)$ & VLAD+PCA \\
		\hline
		{\bf GoogLeNet}   & Pool5 & 1024 & $-$ \\
		\hline
		{\bf GoogLeNet}    & Inception5b & $(7,7,1024)$ & VLAD+PCA \\ 
		\hline
		\multicolumn{4}{|c|}{Scene Recognition}\\
		\hline
		{\bf AlexNet}   & FC7 & 4096 & PM+PCA \\ 
		\hline
		{\bf VGG-VD}   & FC7 & 4096 & PM+PCA \\ 
		\hline 
	\end{tabular}\label{tab:model}
\end{table}
}

Deep image features extracted from the last convolution layer preserve their overall spatial locations or image layouts while the fully-connected CNN layer will lose spatial information. We adopt to encode the last convolutional layer outputs (as feature activation maps) in a form of dense pooling via Fisher Vector (FV)~\cite{Perronnin2010FV} and Vector Locally Aggregated Descriptor (VLAD)~\cite{Jegou2012VLAD}, before feeding them to the fully-connected layer. The dimensions of FV or VLAD encoded deep features are much higher than those of the FC layers. Since there is redundant information from the encoded deep features, Principal Component Analysis (PCA) is performed to reduce the feature dimensionality to 4096 (same to the FC dimension \cite{krizhevsky2012imagenet,Szegedy2014Going}) that makes different encoding schemes more comparable. 

Mined mid-level visual elements based image encoding has proven to be a more discriminative representation in natural scene recognition \cite{Singh2012DiscPat,doersch2013mid,Juneja2013Blocks,LiLSH15CVPR,Wu2015Harvesting}. Visual elements are expected to be common amongst the images with same label but seldom occur in other categories. The association rule mining technique is integrated into our looped optimization method flow (similar to \cite{LiLSH15CVPR}) to automatically discover mid-level image patches for encoding. We conjecture that discriminative patches can be discovered and gradually improved through the LDPO iterations even if the initialization image labels are not accurate.        

{\bf CNN activation based encoding:}
Given a pre-trained (generic or domain-specific) CNN model (e.g., Alexnet or GoogLeNet), an input image $I$ is resized to fit the model definition and feed into the CNN model to extract features $\{f^{L}_{i,j}\}$ ($1\leqslant i,j\leqslant s^{L}$) from the $L$-th convolutional layer with dimensions $s^{L}\times s^{L}\times d^{L}$, e.g., $13\times13\times256$ of $Conv5$ in AlexNet and $7\times7\times1024$ of $Pool5$ in GoogLeNet. For the Fisher Vector implementation, we use the settings as suggested in~\cite{Cimpoi2015Deep}: 64 Gaussian components are adopted to train the Gaussian mixture Model(GMM). The dimension of resulted FV features is significantly higher than $FC7$'s, i.e. $32768 (2\times64\times256) \ vs \ 4096$. After PCA, the FV representation per image is reduced to a $4096$-component vector. A list of deep image features, the encoding methods and output dimensions are provided in Table~\ref{tab:model}. To be consistent with the setting of FV encoding, we initialize the VLAD encoding of convolutional features by $k$-means clustering ($k=64$). Thus the resulted dimensions of VLAD descriptors are $16384(64\times256)$ of $Conv5$ in AlexNet and $65536(64\times1024)$ of $Inception5b$ in GoogLeNet, both reduced to $4096$ via PCA. 

{\bf Patch mining based encoding:}
We adopt a procedure similar to that in \cite{LiLSH15CVPR} to extract mid-level elements for image representation. Our method, however, unlike \cite{LiLSH15CVPR}, does not require prior knowledge of the image categories. For each image $I$ in the dataset, we first extract a set of patches from multiple spatial scales and compute the CNN activation for each patch. Among all activations (e.g., 4096-D vectors on FC7), only indexes of top $k$ maximal activations are recorded and used to form a transaction (e.g., $\{1024, 3, 24, 4096\}$, $k=4$) \cite{LiLSH15CVPR}. Each image contains a set of transactions, which appears on the image. Instead of retrieving patches in a class-specific fashion (\cite{LiLSH15CVPR} with known labels), we employed association rule mining inside the sets of either randomly grouped images (for the first iteration) or image clusters computed by ``clustering on CNN features''. The top 50 mined patterns (which cover the maximum numbers of patches) per image cluster are further merged across the entire dataset to form a consolidated vocabulary of visual elements. Detailed global merging procedures are elaborated in Algorithm \ref{alg-merging}. Compared to \cite{LiLSH15CVPR}, we find that our global merging strategy effectively reduces redundancy and offers more discriminative image features for both clustering and classification tasks (see details in Sec. \ref{sec-Exp}). Finally, the ``bag-of-elements'' image representations are computed as the same process in \cite{LiLSH15CVPR}.

\begin{algorithm}[t]
	\label{alg-merging}
	
	\caption{Global merging of patch clusters}
	\KwIn{A set of mined patterns from each of $K$ image clusters, i.e.$\mathcal{V}=\{v_i\}$, $|\mathcal{V}|$=50$*K$, set of patches $p_i\in \mathcal{P}$ for each pattern and LDA detectors \cite{hariharan2012discriminative} $d_i\in \mathcal{D}$ trained on associated patch set $p_i$.}
	\KwOut{A set of merged patterns $\mathcal{V}'=\{v_n\}$ and associated patch LDA detectors $\mathcal{D}'=\{d_n\}$}
	
	\For{each set of \{$v_i,p_i,d_i$\}}{
		Compute $S_{ij}=1/{|p_j|}\sum_{x \in X_{p_j}}d_i^Tx$ ( $X_{p_j}$ is a set of CNN activations of patches in $p_j$). 
	}
	\If{both $S_{ij}$ and $S_{ji} > \textit{a predefined threshold}$}
	{Merge $\{v_i,p_i,d_i\}$ and $\{v_j,p_j,d_j\}$ and train new LDA detector $d_n$ based on $p_i\cup p_j$. 
	}
	
	\KwRet $\mathcal{V}'$, $\mathcal{D}'$\;
	\BlankLine
	
\end{algorithm}

\subsection{Image Clustering and LDPO Convergence}\label{sec:clustering} 

Image clustering plays an indispensable role in our LDPO framework. We hypothesize that the newly generated clusters driven by looped deep pseudo-task optimization have incrementally improved quality than previous ones, in the following measurements: 1) Images in each cluster are visually more coherent and discriminative from instances in other clusters; 2) The image counts among all clusters are approximately balanced; 3) The number of clusters is self-adaptive by model selection. Two clustering methods are exploited, i.e., standalone $k$-means; or an over-segmented $k$-means (where $K$ is much larger than the first setting, e.g., 1000) followed by RIM~\cite{Gomes2010Discriminative} for model selection and parameter optimization.

$k$-means is an efficient clustering algorithm provided that the number of clusters is known. For scene recognition application, we use $k$-means clustering to initialize the patch mining procedure and generate new image labels for the next iteration, while the underlying cluster number is unknown for the medical image categorization problem. Therefore we first utilize $k$-means clustering to initialize the RIM clustering with a considerably large $k$; then RIM will perform model selection to optimize on $k$. RIM works without the assumption that the cluster number is known as a priori and is designed for discriminative clustering, by maximizing the mutual information between data distribution and the resulted categories via a regularization term on model complexity. The objective function is defined as 
\begin{equation}
f(\mathbf{W};\mathbf{F},\lambda)=I_{\mathbf{W}}\{c;\mathbf{f}\}-R(\mathbf{W};\lambda),
\label{eq:RIM:objective}
\end{equation}
where $c\in\{1,...,K\}$ is a category label, $\mathbf{F}$ is the set of image features $\mathbf{f_{i}}=(f_{i1},...,f_{iD})^{T}\in\mathbb{R}^{D}$. $I_{\mathbf{W}}\{c;\mathbf{f}\}$ is an estimation of the mutual information between the feature vector $\mathbf{f}$ and the label $c$ under the conditional model $p(c|\mathbf{f},\mathbf{W})$. $R(\mathbf{W};\lambda)$ is the complexity penalty and specified according to $p(c|\mathbf{f},\mathbf{W})$. We adopt the unsupervised multilogit regression cost as \cite{Gomes2010Discriminative}. The conditional model and the regularization term are subsequently defined as
\begin{eqnarray}
p(c=k|\mathbf{f},\mathbf{W})&\propto& exp(w^{T}_{k}\mathbf{f}+b_{k}) \\  R(\mathbf{W};\lambda)&=&\lambda\sum_{k}w^{T}_{k}w_{k},
\label{eq:RIM:multilogit}
\end{eqnarray}
where $\mathbf{W}=\{\mathbf{w}_{1},...,\mathbf{w}_{K},b_{1},...,b_{K}\}$ is the set of parameters and $\mathbf{w}_{k}\in\mathbb{R}^{D},b_{k}\in\mathbb{R}$. Maximizing the above objective function is equivalent to solving a logistic regression problem. $R$ is the $L_{2}$ regulator of weight $\{w_{k}\}$ and its power is controlled by $\lambda$.  Large $\lambda$ values enforce reduction of the total number of categories or clusters by imposing no penalty on unpopulated categories ~\cite{Gomes2010Discriminative}. This characteristic enables RIM to attain the optimal number of categories coherent with the data distribution. $\lambda$ is fixed to $1$ in all our experiments.  


Before using the newly-generated clustering labels of image to fine-tune the deep CNN model in the next iteration, the LDPO framework is designed to evaluate the current clustering quality to decide if a convergence has been reached. Two convergence measurements have been adopted from \cite{Tuytelaars09}, i.e., Purity and Normalized Mutual Information (NMI). We take these two criteria as the forms of empirical similarity examination between two clustering outcomes from adjacent LDPO iterations. When the similarity measure is above a certain threshold, we consider that the optimal clustering-based data categorization is reached. It has been empirically found that the final category numbers (from the RIM process) in later LDPO iterations stabilize around a constant. The convergence on classification plots is also observable through the increased top-1, top-5 classification accuracy values in the first few LDPO rounds and eventually stabilize around a constant. 

%
%


{\bf NLP Text Processing:} The category discovery of medical images entails clinically-semantic labeling of the medical images. From the optimized clusters (obtained after Sec. \ref{sec:clustering}), we collect the associated text reports and assemble each image cluster's text reports together into a group. Next, NLP is performed on each unit of radiology reports to find highly recurring words that may serve as informative key words per cluster by counting and ranking the frequency of each word. Common words to all clusters are first removed from the list. The resulting key words and randomly sampled exemplary images for each cluster or category are compiled for reviewing by board-certified radiologists. This process shares some analogy to the human-machine collaborative image database construction \cite{Yu2015Construction,Wigness2015}. 

\subsection{Hierarchical Category Relationship}\label{sec:hcl}

ImageNet \cite{deng2009imagenet} is constructed according to the WordNet ontology hierarchy \cite{Miller1995}. 
In this work, our converged CNN classification model can be further extended to explore the hierarchical class relationship in a tree representation. First, the pairwise class similarity or affinity score $A_{i,j}$ between classes (i,j) is modeled via an adapted measurement of CNN classification confusion. 
{\small
\begin{align}
\small
A_{i,j} &= \frac{1}{2} \Big(Prob(i|j) + Prob(j|i) \Big)  \\
&= \frac{1}{2} \Big(\frac{\sum_{I_m\in C_i}CNN(I_m|j)}{|C_i|} + \frac{\sum_{I_n\in C_j}CNN(I_n|i)}{|C_j|}\Big)
\label{eq:HCL:rij}
\end{align}
}
where $C_i$, $C_j$ are the image sets for class $i$,$j$ respectively, $|\cdot|$ is the cardinality function, $CNN(I_m|j)$ is the CNN classification score of image $I_m$ (from class $C_i$) according to class $j$ that is directly obtained by the N-way CNN softmax. Here $A_{i,j} = A_{j,i}$ is symmetric by averaging $Prob(i|j)$ and $Prob(j|i)$.

The Affinity Propagation algorithm \cite{frey07affinitypropagation} (AP) is then invoked to perform a ``tuning parameter-free'' partitioning on this pairwise affinity matrix $\{A_{i,j}\} \in \mathbb{R}^{K\times K}$. This process can be executed recursively to generate a hierarchically-merged category tree. Without loss of generality, we assume that at level L, classes $i^L$,$j^L$ are formed by merging classes at level L-1 through AP clustering. The new affinity score $A_{i^L,j^L}$ is computed as follows.
{\small
\begin{eqnarray}
\small
A_{i^L,j^L} = \frac{1}{2} \Big(Prob(i^L|j^L) + Prob(j^L|i^L) \Big)
\\  Prob(i^L|j^L) = \frac{\sum_{I_m\in C_{i^L}}\sum_{k \in {j^L}}CNN(I_m|k)}{|C_{i^L}|}
\label{eq:HCL:rijH}
\end{eqnarray}
}
where the L-th level class label ${j^L}$ includes all merged original classes (i.e., 0-th level before AP is called) $k \in {j^L}$ obtained thus far. N-way CNN classification scores only need to be evaluated once at the beginning of AP. The consequent value of any $A_{i^L,j^L}$ at any merged level is the sum of the 0-th level confusion scores. The modeled category hierarchy can alleviate the highly uneven visual separability among discovered image categories \cite{Yan2015hd}. 

\section{Experimental Results}\label{sec-Exp}




{\bf Datasets:}
We experiment on the same medical image dataset as in~\cite{Shin2015} that contains totally 215,786 $2D$ key-images and the associated radiology reports of 61,845 unique patients. Key-images are resized to 256$\times$256 bitmap images (from 512$\times$512). The intensity ranges are rescaled using the default ``optimal'' window settings stored in the DICOM header files (Intensity rescaling improves the CNN classification accuracy by $\sim2\%$ comparing to ~\cite{Shin2015}). Patient-sensitive information in radiology reports is removed for privacy reasons. Furthermore, we quantitatively evaluate our LDPO framework on three widely-reported scene recognition benchmark datasets: 1). I-67 \cite{quattoni2009recognizing} of 67 indoor scene classes with 15620 images; 2). B-25 \cite{xu2014architectural} of 25 architectural styles from 4794 images; 3). S-15 \cite{lazebnik2006beyond} of 15 outdoor and indoor mixed scene classes with 4485 images. For scene recognition, the ground truth (GT) labels are only used to validate the final quantitative LDPO clustering results (where cluster-purity becomes classification accuracy). The cluster number is assumed to be known to LDPO during clustering (Sec. \ref{sec:clustering}) for a fair comparison. Thus the model selection RIM module is dropped.

In each LDPO round, 1) the image clustering step (Sec. \ref{sec:clustering}) is applied on the entire image dataset in order to assign a cluster label to each image, 2) for CNN model fine-tuning (Sec. \ref{sec:ip}), images are randomly reshuffled into three subsets of training ($70\%$), validation ($10\%$) and testing ($20\%$) at each iteration. This ensures that LDPO convergence will generalize to the entire image database. The CNN model is fine-tuned at each LDPO iteration once a new set of image labels is generated from the clustering stage. We use Caffe~\cite{jia2014caffe} implementation of CNN models. The softmax loss layer (i.e., 'FC8' in AlexNet and 'loss3/classifier' in GoogLeNet) is more significantly modulated by 1) setting a higher learning rate than all other CNN layers; and 2) updating the (varying but converging) number of category classes from the clustering results. 

\begin{figure}[!t]
	\begin{center}
		\includegraphics[width=0.282\linewidth]{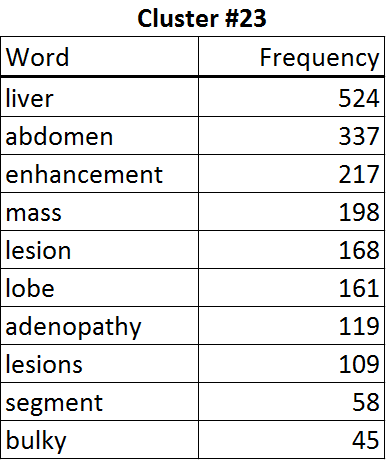}
		\includegraphics[width=0.34\linewidth]{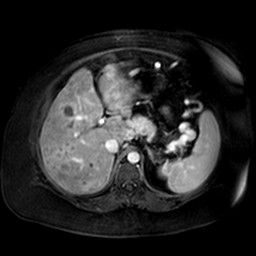}
		\includegraphics[width=0.34\linewidth]{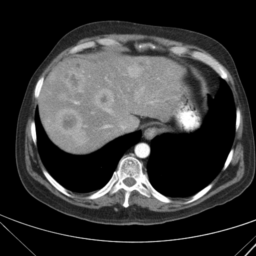}
		\includegraphics[width=0.282\linewidth]{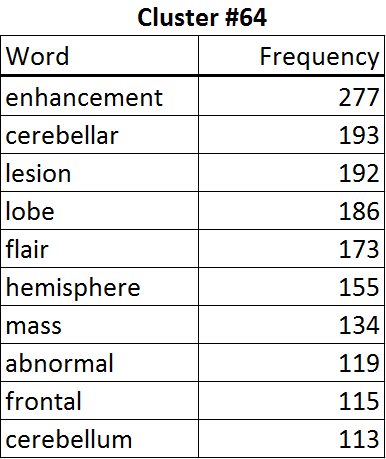}
		\includegraphics[width=0.34\linewidth]{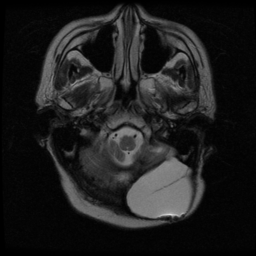}
		\includegraphics[width=0.34\linewidth]{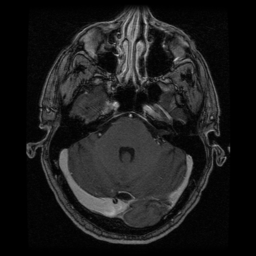}
	\end{center}   
	\caption{Sample images of two unsupervisedly discovered image clusters with associated clinically semantic key words, containing (likely appeared) anatomies, pathologies , their attributes and imaging protocols or properties.}  
	\label{fig:Cluster_Sample:png}
\end{figure}

\begin{figure*}[t]
	\begin{center}
		
		\begin{tabular}{cccc}
			\includegraphics[width=0.235\linewidth]{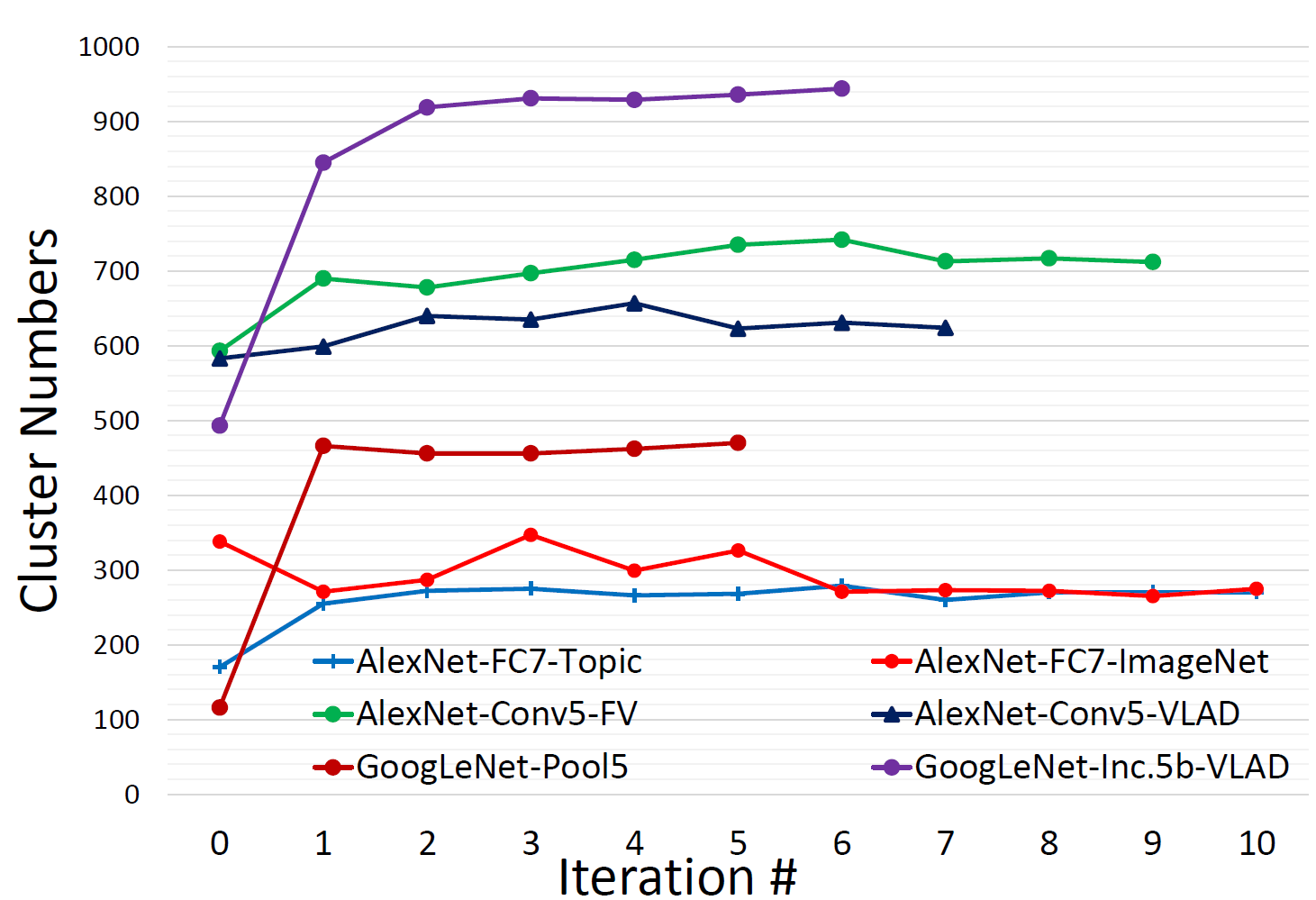} &
			\includegraphics[width=0.235\linewidth]{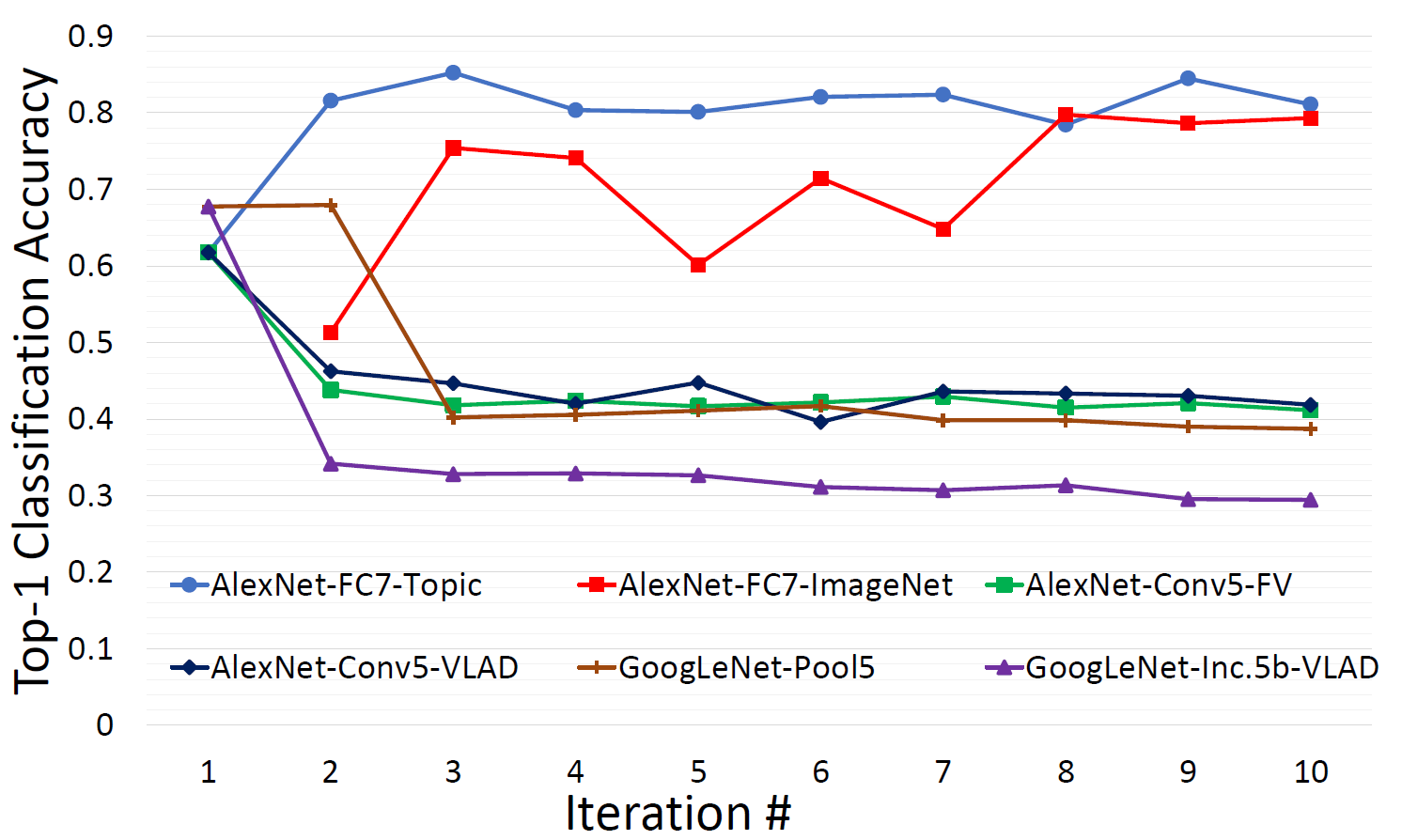} &
			\includegraphics[width=0.235\linewidth]{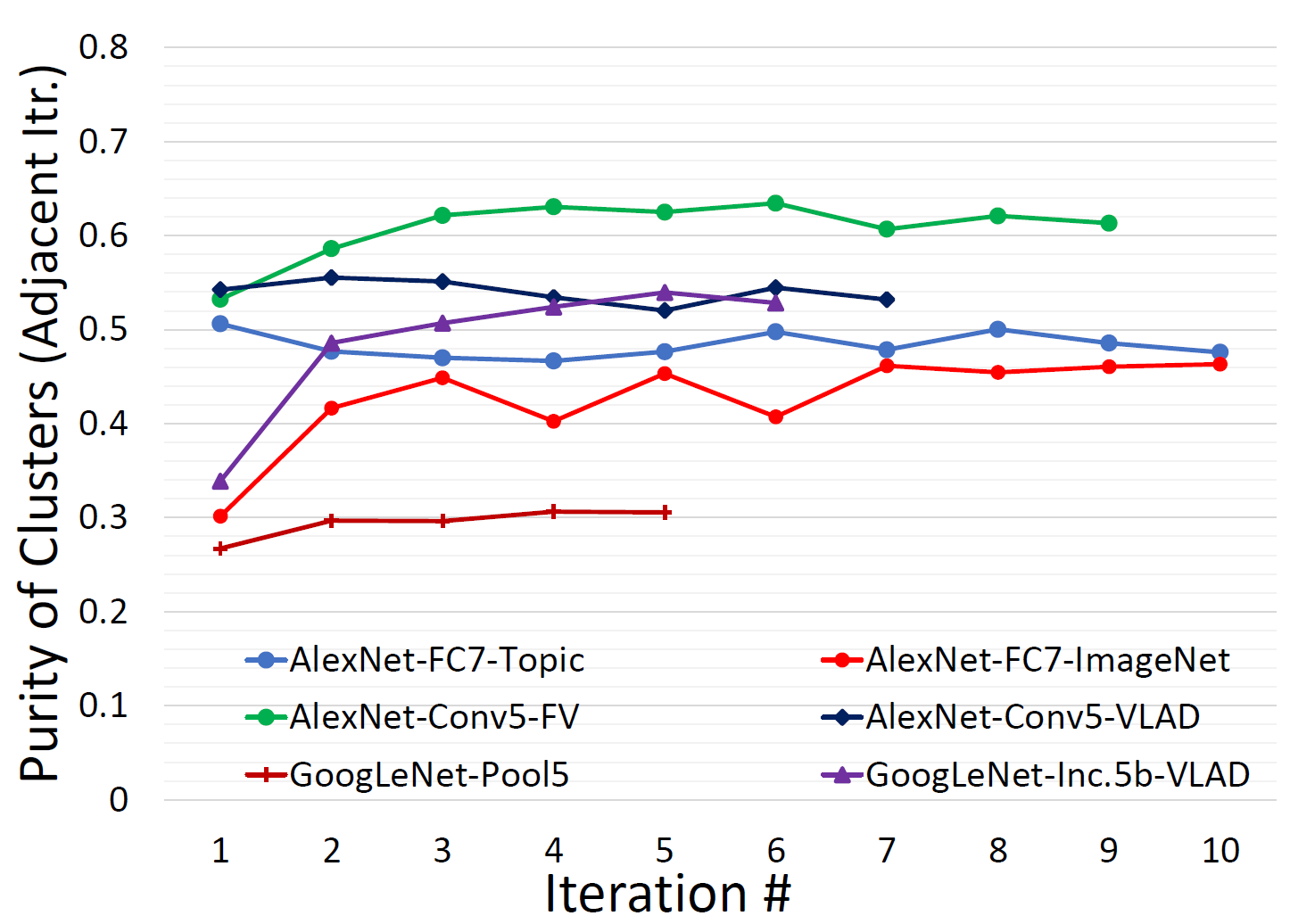} &
			\includegraphics[width=0.235\linewidth]{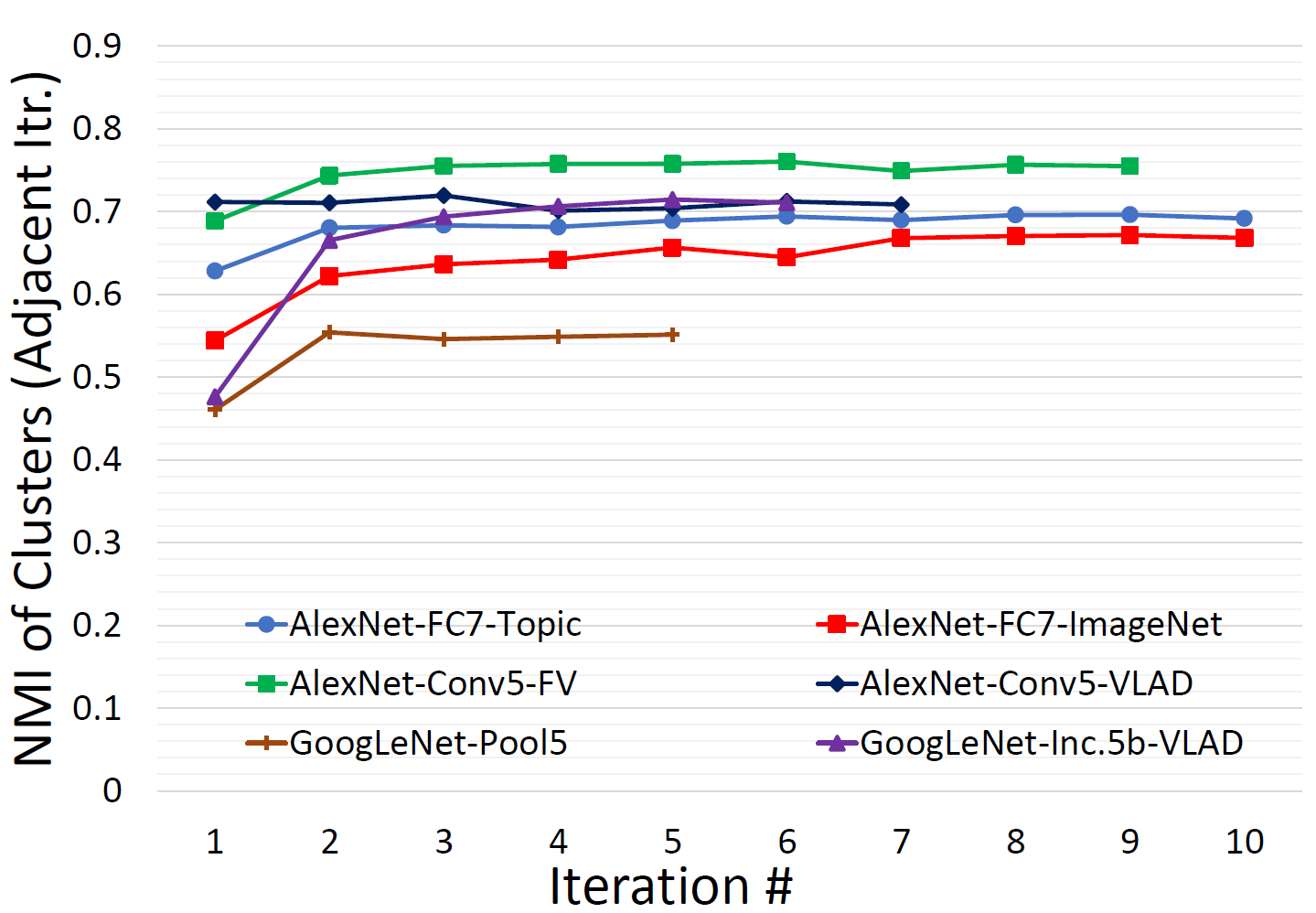}		\\
			(a) & (b) & (c)  & (d)  \\	
		\end{tabular}
		
	\end{center}
	\caption{LDPO Performance using RIM clustering under different image encoding methods (\textit{i.e.}, FV and VLAD) and CNN Architectures (\textit{i.e.}, AlexNet and GoogLeNet). The number of clusters discovered, Top-1 accuracy of trained CNNs, the purity and NMI measurements of clusters from adjacent iterations are illustrated in (a, b, c, d), respectively.}
	\label{fig:RIM_Encode:png}
\end{figure*}

\subsection{Unsupervised Medical Image Categorization}\label{sec-label-results} 
We first investigate the convergence issue of the LDPO method under different system configurations and then report the CNN classification performance on the discovered categories. 

{\small\small\small
	\begin{table}[t!]
		\small
		\caption{Classification Accuracy of Converged CNN Models}
		\centering
		\begin{tabular}{|l||c|c|c|}
			\hline\hline
			{\bf CNN setting} & {\bf Cluster \#} & {\bf Top-1} & {\bf Top-5} \\[0.3ex]
			\hline\hline
			{\bf AlexNet-FC7-Topic}    & 270 & 0.8109 & 0.9412\\[0.3ex]
			\hline
			{\bf AlexNet-FC7-ImageNet}    & 275 & 0.8099 & 0.9547 \\[0.3ex]
			\hline
			{\bf AlexNet-Conv5-FV}   & 712  & 0.4115 & 0.4789  \\[0.3ex]
			\hline
			{\bf AlexNet-Conv5-VLAD}   & 624 & 0.4333 & 0.5232 \\[0.3ex]
			\hline
			{\bf GoogLeNet-Pool5}    & 462 & 0.4109 & 0.5609  \\[0.3ex]
			\hline
			{\bf GoogLeNet-Inc.5b-VLAD}   & 929 & 0.3265 & 0.4001  \\[0.3ex]
			\hline 
		\end{tabular}\label{tab:CNN-Acc}
	\end{table}
}

{\bf Clustering Method:}
As shown in Fig.~\ref{fig:RIM_Encode:png} (a), RIM can estimate unsupervised category numbers consistently well under different image representations (deep CNN feature configurations + encoding schemes). Standalone $k$-means clustering enables LDPO to converge quickly with high classification accuracies whereas RIM based model selection module produces more balanced and semantically meaningful clustering results (see more in Sec.~\ref{sec-label-results}). This advantage is probably due to RIM's two unique properties: 1) less restricted geometric assumptions in the clustering feature space; 
2) the capacity to attain the optimal number of clusters by maximizing the mutual information between input data and the induced clusters via a regularized term. 

{\small
	\begin{table*}[t]
		\centering
		\caption{Clustering performance of LOM and other methods on 3 scene recognition datasets. The last Column presents the state-of-the-art fully-supervised scene Classification Accuracy (CA)  for each dataset, produced by \cite{Cimpoi2015Deep,peng2015framework,Zhou2014Place} respectively.}               
		\begin{tabular}{|l||c|c|c|c|c|c|c|c||c|}
			\hline\hline \small 
			{\bf Dataset}   & {\bf \footnotesize KM \cite{vedaldi08vlfeat}} & {\bf \footnotesize LSC \cite{chen2011large}} & {\bf AC \footnotesize \cite{gowda1978agglomerative}} & {\bf EP \footnotesize \cite{Dai2013EnPro}} & {\bf \footnotesize MDPM \cite{LiLSH15CVPR}} & {\bf \footnotesize LDPO-A-FC} & {\bf \footnotesize LDPO-A-PM} & {\bf \footnotesize LDPO-V-PM} & {\bf \footnotesize Supervised} \\
			\hline\hline
			&\multicolumn{8}{|c||}{Clustering Accuracy (\%)} & CA (\%)\\
			\hline
			{\bf I-67 \cite{quattoni2009recognizing}}   & 35.6 & 30.3 & 34.6 & 37.2 & 53.0 & 37.9 & 63.2 &\textbf{75.3} & \textbf{81.0} \cite{Cimpoi2015Deep}\\ 
			\hline
			{\bf B-25 \cite{xu2014architectural}}    & 42.2 & 42.6 & 43.2 & 43.8 & 43.1 & 44.2 & 59.2 & \textbf{59.5} & \textbf{59.1} \cite{peng2015framework}\\
			\hline
			{\bf S-15 \cite{lazebnik2006beyond}}   & 65.0 & 76.5 & 65.2 & 73.6 & 63.4 & 73.1 & \textbf{90.2} & 84.0 & \textbf{91.6} \cite{Zhou2014Place} \\
			\hline
			&\multicolumn{8}{|c||}{Normalized Mutual Information} & \\
			\hline
			{\bf I-67 \cite{quattoni2009recognizing}}   & .386 & .335 & .359 & - &.558 & .389 & .621 & \textbf{.759} & -\\ 
			\hline
			{\bf B-25 \cite{xu2014architectural}}   & .401 & .403 & .404 & - & .424 & .407 & \textbf{.588} & .546 & -\\ 
			\hline
			{\bf S-15 \cite{lazebnik2006beyond}}   & .659 & .625 & .653 & - & .596 & .705 & \textbf{.861} & .831 & - \\ 
			\hline
		\end{tabular}\label{tab:clustering}
	\end{table*}
}

{\bf Pseudo-Task Initialization:}
Both generic and domain-specific CNN models ~\cite{krizhevsky2012imagenet,Szegedy2014Going,Shin2015} are employed for LDPO initialization. Fig.~\ref{fig:RIM_Encode:png} illustrates the performance of LDPO using two CNN variants -- AlexNet-FC7-ImageNet and AlexNet-FC7-Topic. AlexNet-FC7-ImageNet yields noticeably slower LDPO convergence than its counterpart of AlexNet-FC7-Topic, as the latter has already been fine-tuned by the report-derived category information on the same radiology image database \cite{Shin2015}.. Nevertheless, the final clustering outcomes are similar after convergence from AlexNet-FC7-ImageNet or AlexNet-FC7-Topic. At $\sim10$ iterations, two different initializations result in similar cluster numbers, purity/NMI scores and even classification accuracies (Table~\ref{tab:CNN-Acc}). 

{\bf Deep CNN Feature and Image Encoding:}
Different configurations of image representation can affect the performance of medical image categorization, as shown in Fig.~\ref{fig:RIM_Encode:png}. Deep images features are extracted at different layers of depth from two CNN models (i.e., AlexNet, GoogLeNet) and may present the depth-specific visual information. 
Different image feature encoding schemes (FV or VLAD) add further options or variations into this process. The numbers of clusters range from 270 ({\bf AlexNet-FC7-Topic} with no explicit feature encoding scheme) to 931 (the more sophisticated {\bf GoogLeNet-Inc.5b-VLAD} with VLAD encoding). The numbers of clusters discovered by RIM are expected to reflect the amount of knowledge or information complexity stored in the PACS database. 


{\bf Unsupervised Categorization:} 
Our category discovery clusters are generally visually coherent within the cluster and size-balanced across clusters. However, image clusters formed only based on text information (of $\sim780K$ radiology reports) are highly unbalanced \cite{Shin2015}, with three clusters inhabiting the majority of images. Note that our method imposes no explicit constraint on the number of instances per cluster. 
Fig.~\ref{fig:Cluster_Sample:png} shows sample images and their top-10 associated key words from two randomly selected clusters (more results are provided in the supplementary material). The LDPO clusters are found to be clinically or semantically related to the corresponding key words, which describe presented anatomies, pathologies (e.g., adenopathy, mass), their associated attributes (e.g., bulky, frontal) and imaging protocols or properties.



{\bf Categorization Recognizable?} We validate the following hypothesis: a high quality unsupervised image categorization scheme will generate labels that can be more easily recognized by any supervised CNN model. From Table~\ref{tab:CNN-Acc}, {\bf AlexNet-FC7-Topic} has the {\bf Top-1} classification accuracy of 0.8109 and {\bf Top-5} accuracy 0.9412 with 270 formed image categories while {\bf AlexNet-FC7-ImageNet} achieves the accuracies of 0.8099 and 0.9547, from 275 discovered classes. In contrast, \cite{Shin2015} reports the {\bf Top-1} accuracies of 0.6072, 0.6582 and {\bf Top-5} as 0.9294, 0.9460 from only 80 classes using AlexNet \cite{krizhevsky2012imagenet} or VGGNet-19 \cite{simonyan2014very}, respectively. The classification accuracies shown in Table 2 are computed using the final LDPO-converged CNN models and the testing dataset. Markedly better accuracies (especially on {\bf Top-1}) on classifying higher numbers of classes ( that are generally more challenging) also demonstrate the advantages of the LDPO discovered image clusters or labels over those in \cite{Shin2015}, under the same radiology database. Upon evaluation by two board-certified radiologists, {\bf AlexNet-FC7-Topic} of 270 categories and {\bf AlexNet-FC7-ImageNet} of 275 classes are considered the best of total six model-feature-encoding setups. Interestingly, both models have no external feature encoding schemes built-in and preserve gloss image layouts (without spatially unordered FV or VLAD encoding modules \cite{Cimpoi2015Deep,Jegou2012VLAD}). Refer to supplementary material for more results on radiologists' evaluation.  

\subsection{Scene Recognition}\label{sec-scene-results}
We use three scene recognition datasets to quantitatively evaluate the proposed LDPO-PM method (with patch mining) based on two metrics: 1) clustering based scene recognition accuracy and 2) supervised classification (e.g., Liblinear ) on image representations learned in an unsupervised fashion. 
The purity and NMI measurements are computed between the final LDPO clusters and GT scene classes where {\bf purity becomes the classification accuracy against GT}. The LDPO cluster numbers are set to match the GT class numbers of (67, 25, 15), respectively. We compare the LDPO scene recognition performance to those of several popular clustering methods, such as {\bf KM} \cite{vedaldi08vlfeat}: $k$-means; {\bf LSC} \cite{chen2011large}; {\bf AC} \cite{gowda1978agglomerative}: Agglomerative clustering; {\bf EP} \cite{Dai2013EnPro}: Ensemble Projection $+$ kmeans; and {\bf MDPM} \cite{LiLSH15CVPR}: Mid-level Discriminative Patch Mining $+$ kmeans. Both EP and MDPM use mid-level visual elements based image representations. Three variants of our method (i.e., {\bf LDPO-A-FC7}: FC7 feature on AlexNet, {\bf LDPO-A-PM}: FC7 feature on AlexNet with patch mining, and {\bf LDPO-V-PM}: FC7 feature on VGG-VD with patch mining) are exploited. On all three datasets, the {\bf LDPO-A-PM} and {\bf LDPO-V-PM} achieve significantly higher purity and NMI values than the previous clustering methods (cf. Table \ref{tab:clustering}). Especially for the MIT-67 indoor scene dataset \cite{quattoni2009recognizing}, our best model {\bf LDPO-V-PM} achieves the unsupervised scene recognition accuracy of 75.3\%, which nearly doubles the performances of KM and AC on FC7 features of an ImageNet pretrained AlexNet \cite{krizhevsky2012imagenet,Chatfield14}. Note that the state-of-the-art supervised classification accuracy on MIT-67 is 81.0\%~\cite{Cimpoi2015Deep} and our unsupervised method is comparatively close to that. VGG-VD -- a deeper CNN model -- empirically boosts the recognition performance from {\bf LDPO-A-PM} of 63.2\% to {\bf LDPO-V-PM} at 75.3\% on MIT-67. However this performance gain is not observed on two other smaller datasets.

Next, we evaluate the supervised discriminative power of LDPO-PM learned image representation. We measure its classification accuracy using the MIT-67 dataset and its standard partition \cite{quattoni2009recognizing}, i.e., 80 training and 20 testing images per class. As in \cite{Singh2012DiscPat,sun2013learning,doersch2013mid,LiLSH15CVPR,Wu2015Harvesting,Cimpoi2015Deep}, we use the Liblinear classification toolbox \cite{fan2008liblinear} on the {\bf LDPO-V-PM} image representation (noted as {\bf LDPO-V-PM-LL}), under 5-fold cross validation. The supervised and unsupervised scene recognition accuracy results from previous state-of-the-art work and variants of our method are listed in Table  \ref{tab:classification}. The one-versus-all Liblinear classification in {\bf LDPO-V-PM-LL} does not noticeably improve upon purely unsupervised {\bf LDPO-V-PM}. This may indicate that the LDPO-PM image representation is sufficient to adequately separate images from different scene classes. Last, we experiment the clustering convergence issue with two different initializations: random initialization or image labels obtained from k-means clustering on FC7 features of an ImageNet pretrained AlexNet. 
While the clustering accuracy of the LDPO-PM with random initialization increases rapidly during its first iterations, both schemes ultimately converge to similar performance levels. This suggests that the LDPO convergence is insensitive to the chosen initialization.

{\small\small\small
	\begin{table}[t!]
		\caption{Scene Recognition Accuracy on MIT-67 Dataset \cite{quattoni2009recognizing}}
		\centering
		\begin{tabular}{|l||c|}
			\hline\hline
			{\bf Method} & {\bf Accuracy (\%)} \\[0.3ex]
			\hline\hline
			{\bf D-patch \cite{Singh2012DiscPat} }    & 38.1  \\[0.3ex]
			\hline
			{\bf D-parts \cite{sun2013learning} }    & 51.4  \\[0.3ex]
			\hline
			{\bf DMS \cite{doersch2013mid} }   & 64.0 \\[0.3ex]
			\hline
			{\bf MDPM-VGG \cite{LiLSH15CVPR} }    & 77.0 \\[0.3ex]
			\hline
			{\bf MetaObject \cite{Wu2015Harvesting} }   & 78.9 \\[0.3ex]
			\hline 
			\hline
			{\bf FC (VGG) }   & 68.9  \\[0.3ex]
			\hline
			{\bf CONV-FV (VGG) \cite{Cimpoi2015Deep}}   & {\bf 81.0} \\[0.3ex]
			\hline 
			\hline
			{\bf LDPO-V-PM-LL}   & 72.5  \\[0.3ex]
			\hline
			{\bf LDPO-V-PM}   & {\bf 75.3}  \\[0.3ex]
			\hline 
						
		\end{tabular}\label{tab:classification}
	\end{table}
}

\section{Conclusion} 

In this paper, we present a Looped Deep Pseudo-task Optimization framework for unsupervised joint mining of image features and labels. Our method is  validated through two important problems: 1) discovery and exploration of semantic categories from a large-scale medical image database and 2) unsupervised scene cognition on three public datasets. Extensive experiments demonstrate excellent quantitative and qualitative results on both tasks. The measurable LDPO ``convergence'' makes the ill-posed image auto-annotation problem better constrained. 

{ {\bf Acknowledgements  } This work was supported by the Intramural Research Program of the NIH Clinical Center. This work utilized the computational resources of the NIH HPC Biowulf cluster (http://hpc.nih.gov). We thank NVIDIA Corporation for the GPU donation.}

{\small
\bibliographystyle{ieee}
\bibliography{LDPO_WACV2017}

\begin{thebibliography}{10}\itemsep=-1pt

\bibitem{blei2003latent}
D.~M. Blei, A.~Y. Ng, and M.~I. Jordan.
\newblock Latent dirichlet allocation.
\newblock {\em Journal of machine Learning research}, 3:993--1022, 2003.

\bibitem{bossard2014food}
L.~Bossard, M.~Guillaumin, and L.~Van~Gool.
\newblock Food-101--mining discriminative components with random forests.
\newblock In {\em European Conference on Computer Vision}, pages 446--461.
  Springer, 2014.

\bibitem{Chatfield14}
K.~Chatfield, K.~Simonyan, A.~Vedaldi, and A.~Zisserman.
\newblock Return of the devil in the details: Delving deep into convolutional
  nets.
\newblock In {\em British Machine Vision Conference}, 2014.

\bibitem{chen2011large}
X.~Chen and D.~Cai.
\newblock Large scale spectral clustering with landmark-based representation.
\newblock In {\em AAAI}, 2011.

\bibitem{Chen2015Webly}
X.~Chen and A.~Gupta.
\newblock Webly supervised learning of convolutional networks.
\newblock In {\em Proc. of ICCV}, 2015.

\bibitem{Chen2013Neil}
X.~Chen, A.~Shrivastava, and A.~Gupta.
\newblock Neil: Extracting visual knowledge from web data.
\newblock In {\em Proc. of ICCV}, 2013.

\bibitem{choi2013adding}
J.~Choi, M.~Rastegari, A.~Farhadi, and L.~S. Davis.
\newblock Adding unlabeled samples to categories by learned attributes.
\newblock In {\em Proceedings of the IEEE Conference on Computer Vision and
  Pattern Recognition}, pages 875--882, 2013.

\bibitem{Cimpoi2015Deep}
M.~Cimpoi, S.~Maji, and A.~Vedaldi.
\newblock Deep filter banks for texture recognition and segmentation.
\newblock {\em Proc. of IEEE CVPR}, 2015.

\bibitem{Coates2011}
A.~Coates, A.~Ng, and H.~Lee.
\newblock An analysis of single-layer networks in unsupervised feature
  learning.
\newblock {\em AI and Statistics}, 2011.

\bibitem{Dai2013EnPro}
D.~Dai and L.~{Van Gool}.
\newblock Ensemble projection for semi-supervised image classification.
\newblock In {\em Proc. of ICCV}, 2013.

\bibitem{Dai2015EnProDeepFets}
D.~Dai and L.~{Van Gool}.
\newblock Unsupervised high-level feature learning by ensemble projection for
  semi-supervised image classification and image clustering.
\newblock Technical report, arXiv:1602.00955, 2016.

\bibitem{deng2009imagenet}
J.~Deng, W.~Dong, R.~Socher, L.-J. Li, K.~Li, and L.~Fei-Fei.
\newblock Imagenet: A large-scale hierarchical image database.
\newblock In {\em Computer Vision and Pattern Recognition}, pages 248--255.
  IEEE, 2009.

\bibitem{doersch2013mid}
C.~Doersch, A.~Gupta, and A.~A. Efros.
\newblock Mid-level visual element discovery as discriminative mode seeking.
\newblock In {\em Advances in Neural Information Processing Systems (NIPS)},
  pages 494--502, 2013.

\bibitem{doersch2015unsupervised}
C.~Doersch, A.~Gupta, and A.~A. Efros.
\newblock Unsupervised visual representation learning by context prediction.
\newblock In {\em Proceedings of the IEEE International Conference on Computer
  Vision}, pages 1422--1430, 2015.

\bibitem{Dosovitskiy2014}
A.~Dosovitskiy, J.~Springenberg, M.~Riedmiller, and T.~Brox.
\newblock Discriminative unsupervised feature learning with convolutional
  neural networks.
\newblock {\em NIPS}, 2014.

\bibitem{Everingham2015Pascal}
M.~Everingham, S.~M.~A. Eslami, L.~Van~Gool, C.~Williams, J.~Winn, and
  A.~Zisserman.
\newblock The pascal visual object classes challenge: A retrospective.
\newblock {\em International journal of computer vision}, 111(1):98--136, 2015.

\bibitem{fan2008liblinear}
R.-E. Fan, K.-W. Chang, C.-J. Hsieh, X.-R. Wang, and C.-J. Lin.
\newblock Liblinear: A library for large linear classification.
\newblock {\em Journal of machine learning research}, 9(Aug):1871--1874, 2008.

\bibitem{FeiFei2004101}
L.~Fei-Fei, R.~Fergus, and P.~Perona.
\newblock Learning generative visual models from few training examples: an
  incremental bayesian approach tested on 101 object categories.
\newblock {\em Proc. of IEEE CVPR workshop}, 2004.

\bibitem{frey07affinitypropagation}
B.~Frey and D.~Dueck.
\newblock Clustering by passing messages between data points.
\newblock {\em Science}, 315:972--976, 2007.

\bibitem{Girshick2015RCNN}
R.~Girshick, J.~Donahue, T.~Darrell, and J.~Malik.
\newblock Region-based convolutional networks for accurate object detection and
  semantic segmentation.
\newblock {\em IEEE Trans. Pattern Anal. Mach. Intell.}, 2015.

\bibitem{Gomes2010Discriminative}
R.~Gomes, A.~Krause, and P.~Perona.
\newblock Discriminative clustering by regularized information maximization.
\newblock {\em NIPS}, 2010.

\bibitem{gowda1978agglomerative}
K.~C. Gowda and G.~Krishna.
\newblock Agglomerative clustering using the concept of mutual nearest
  neighbourhood.
\newblock {\em Pattern recognition}, 10(2):105--112, 1978.

\bibitem{hariharan2012discriminative}
B.~Hariharan, J.~Malik, and D.~Ramanan.
\newblock Discriminative decorrelation for clustering and classification.
\newblock In {\em European Conference on Computer Vision}, pages 459--472.
  Springer, 2012.

\bibitem{Huh2016what}
M.~Huh, P.~Agrawal, and A.~A. Efros.
\newblock What makes imagenet good for transfer learning?
\newblock In {\em arXiv preprint: arXiv:1608.08614}, 2016.

\bibitem{Jegou2012VLAD}
H.~Jegou, F.~Perronnin, M.~Douze, J.~Sanchez, P.~Perez, and C.~Schmid.
\newblock Aggregating local image descriptors into compact codes.
\newblock {\em Pattern Analysis and Machine Intelligence, IEEE Transactions
  on}, 34(9):1704--1716, Sept 2012.

\bibitem{jia2014caffe}
Y.~Jia, E.~Shelhamer, J.~Donahue, S.~Karayev, J.~Long, R.~Girshick,
  S.~Guadarrama, and T.~Darrell.
\newblock Caffe: Convolutional architecture for fast feature embedding.
\newblock {\em arXiv preprint arXiv:1408.5093}, 2014.

\bibitem{Juneja2013Blocks}
M.~Juneja, A.~Vedaldi, C.~Jawahar, and A.~Zisserman.
\newblock Blocks that shout: Distinctive parts for scene classification.
\newblock {\em CVPR}, pages 923--930, 2013.

\bibitem{Karpathy2015Deep}
A.~Karpathy and L.~Fei-Fei.
\newblock Deep visual-semantic alignments for generating image descriptions.
\newblock {\em Proc. of IEEE CVPR}, pages 3128--3137, 2015.

\bibitem{Kingma2014SSL}
D.~Kingma, S.~Mohamed, D.~Rezende, and M.~Welling.
\newblock Semi-supervised learning with deep generative models.
\newblock {\em NIPS}, 2014.

\bibitem{krizhevsky2012imagenet}
A.~Krizhevsky, I.~Sutskever, and G.~E. Hinton.
\newblock Imagenet classification with deep convolutional neural networks.
\newblock In {\em Advances in neural information processing systems}, pages
  1097--1105, 2012.

\bibitem{Lazebnik2005Sparse}
S.~Lazebnik, C.~Schmid, and J.~Ponce.
\newblock A sparse texture representation using local affine regions.
\newblock {\em IEEE Trans. Pattern Anal. Mach. Intell.}, 27(8):1265--1278,
  2005.

\bibitem{lazebnik2006beyond}
S.~Lazebnik, C.~Schmid, and J.~Ponce.
\newblock Beyond bags of features: Spatial pyramid matching for recognizing
  natural scene categories.
\newblock In {\em 2006 IEEE Computer Society Conference on Computer Vision and
  Pattern Recognition (CVPR'06)}, volume~2, pages 2169--2178. IEEE, 2006.

\bibitem{li2013harvesting}
Q.~Li, J.~Wu, and Z.~Tu.
\newblock Harvesting mid-level visual concepts from large-scale internet
  images.
\newblock In {\em Proceedings of the IEEE Conference on Computer Vision and
  Pattern Recognition}, pages 851--858, 2013.

\bibitem{LiLSH15CVPR}
Y.~Li, L.~Liu, C.~Shen, and A.~van~den Hengel.
\newblock Mid-level deep pattern mining.
\newblock In {\em CVPR}, pages 971--980, 2015.

\bibitem{Li2011Towards}
Y.~Li and Z.~Zhou.
\newblock Towards making unlabeled data never hurt.
\newblock {\em ICML}, 2011.

\bibitem{Liang2015Baby}
X.~Liang, S.~Liu, Y.~Wei, L.~Liu, L.~Lin, and S.~Yan.
\newblock Computational baby learning.
\newblock In {\em Proc. of ICCV}, 2015.

\bibitem{Lin2015Nin}
M.~Lin, Q.~Chen, and S.~Yan.
\newblock Network in network.
\newblock In {\em Proc. of ICLR}, 2015.

\bibitem{Lin2014Coco}
T.~Lin, M.~Maire, S.~Belongie, J.~Hays, P.~Perona, D.~Ramanan, P.~Dollar, and
  L.~Zitnick.
\newblock Microsoft coco: Common objects in context.
\newblock In {\em ECCV}, 2014.

\bibitem{Liu2010Large}
W.~Liu, J.~He, and S.~Chang.
\newblock Large graph construction for scalable semi-supervised learning.
\newblock {\em ICML}, 2010.

\bibitem{Miller1995}
G.~A. Miller.
\newblock Wordnet: a lexical database for english.
\newblock {\em Communications of the ACM}, 1995.

\bibitem{Ng15}
J.~Y. Ng, F.~Yang, and L.~S. Davis.
\newblock Exploiting local features from deep networks for image retrieval.
\newblock {\em CoRR}, abs/1504.05133, 2015.

\bibitem{peng2015framework}
K.-C. Peng and T.~Chen.
\newblock A framework of extracting multi-scale features using multiple
  convolutional neural networks.
\newblock In {\em 2015 IEEE International Conference on Multimedia and Expo
  (ICME)}, pages 1--6. IEEE, 2015.

\bibitem{Perronnin2010FV}
F.~Perronnin, J.~Sánchez, and T.~Mensink.
\newblock Improving the fisher kernel for large-scale image classification.
\newblock In {\em Computer Vision – ECCV 2010}, volume 6314 of {\em Lecture
  Notes in Computer Science}, pages 143--156. Springer Berlin Heidelberg, 2010.

\bibitem{quattoni2009recognizing}
A.~Quattoni and A.~Torralba.
\newblock Recognizing indoor scenes.
\newblock In {\em Computer Vision and Pattern Recognition, IEEE Conference on},
  pages 413--420. IEEE, 2009.

\bibitem{Quattoni2009indoor}
A.~Quattoni and A.~Torralba.
\newblock Recognizing indoor scenes.
\newblock {\em Proc. of IEEE CVPR}, 2009.

\bibitem{Raina2007Self}
R.~Raina, A.~Battle, H.~Lee, B.~Packer, and A.~Ng.
\newblock Self-taught learning: transfer learning from unlabeled data.
\newblock {\em ICML}, 2007.

\bibitem{Razavian2014CNN}
A.~Razavian, H.~Azizpour, J.~Sullivan, and S.~Carlsson.
\newblock Cnn features off-the-shelf: an astounding baseline for recognition.
\newblock {\em ArXiv:1403.6382}, 2014.

\bibitem{russakovsky2014imagenet}
O.~Russakovsky, J.~Deng, H.~Su, J.~Krause, S.~Satheesh, S.~Ma, Z.~Huang,
  A.~Karpathy, A.~Khosla, M.~Bernstein, et~al.
\newblock Imagenet large scale visual recognition challenge.
\newblock {\em arXiv preprint arXiv:1409.0575}, 2014.

\bibitem{Shin2015}
H.~Shin, L.~Lu, L.~Kim, A.~Seff, J.~Yao, and R.~Summers.
\newblock Interleaved text/image deep mining on a large-scale radiology
  database.
\newblock {\em Proc. of IEEE CVPR}, 2015.

\bibitem{Shin2015Interleaved}
H.~Shin, L.~Lu, L.~Kim, A.~Seff, J.~Yao, and R.~Summers.
\newblock Interleaved text/image deep mining on a large-scale radiology image
  database for automated image interpretation.
\newblock {\em Journal of Machine Learning Research}, pages 17(107): 1--31,
  2016.

\bibitem{shrivastava2012constrained}
A.~Shrivastava, S.~Singh, and A.~Gupta.
\newblock Constrained semi-supervised learning using attributes and comparative
  attributes.
\newblock In {\em European Conference on Computer Vision}, pages 369--383.
  Springer, 2012.

\bibitem{simonyan2014very}
K.~Simonyan and A.~Zisserman.
\newblock Very deep convolutional networks for large-scale image recognition.
\newblock In {\em Proc. Int. Conf. Learn. Repr.}, 2015.

\bibitem{Singh2012DiscPat}
S.~Singh, A.~Gupta, and A.~A. Efros.
\newblock Unsupervised discovery of mid-level discriminative patches.
\newblock In {\em European Conference on Computer Vision}, 2012.

\bibitem{sun2013learning}
J.~Sun and J.~Ponce.
\newblock Learning discriminative part detectors for image classification and
  cosegmentation.
\newblock In {\em Proceedings of the IEEE International Conference on Computer
  Vision}, pages 3400--3407, 2013.

\bibitem{Szegedy2014Going}
C.~Szegedy, W.~Liu, Y.~Jia, P.~Sermanet, S.~Reed, D.~Anguelov, D.~Erhan,
  V.~Vanhoucke, and A.~Rabinovich.
\newblock Going deeper with convolutions.
\newblock {\em IEEE Conf. on Computer Vision and Pattern Recognition,
  arXiv:1409.4842}, 2015.

\bibitem{Tuytelaars09}
T.~Tuytelaars, C.~H. Lampert, M.~B. Blaschko, and W.~Buntine.
\newblock Unsupervised object discovery: {A} comparison.
\newblock {\em International Journal of Computer Vision}, 2009.

\bibitem{vedaldi08vlfeat}
A.~Vedaldi and B.~Fulkerson.
\newblock {VLFeat}: An open and portable library of computer vision algorithms,
  2008.

\bibitem{Wigness2015}
M.~Wigness, B.~Draper, and J.~Beveridge.
\newblock Efficient label collection for unlabeled image datasets.
\newblock {\em Proc. of IEEE CVPR}, 2015.

\bibitem{Wu2015Deep}
J.~Wu, Y.~Yu, C.~Huang, and K.~Yu.
\newblock Deep multiple instance learning for image classification and
  auto-annotation.
\newblock {\em Proc. of CVPR}, pages 3460--3469, 2015.

\bibitem{Wu2015Harvesting}
R.~Wu, B.~Wang, and Y.~Yu.
\newblock Harvesting discriminative meta objects with deep cnn features for
  scene classification.
\newblock In {\em Proc. of ICCV}, 2015.

\bibitem{Xiao2010Sun}
J.~Xiao, J.~Hays, K.~Ehinger, A.~Oliva, and A.~Torralba.
\newblock Sun database: Large-scale scene recognition from abbey to zoo.
\newblock In {\em CVPR}, pages 3485--3492, 2010.

\bibitem{xu2014architectural}
Z.~Xu, D.~Tao, Y.~Zhang, J.~Wu, and A.~C. Tsoi.
\newblock Architectural style classification using multinomial latent logistic
  regression.
\newblock In {\em European Conference on Computer Vision}, pages 600--615.
  Springer, 2014.

\bibitem{Yan2015hd}
Z.~Yan, H.~Zhang, R.~Piramuthu, V.~Jagadeesh, D.~DeCoste, W.~Di, and Y.~Yu.
\newblock Hd-cnn: Hierarchical deep convolutional neural network for large
  scale visual recognition.
\newblock {\em Proc. of ICCV}, 2015.

\bibitem{yang2016joint}
J.~Yang, D.~Parikh, and D.~Batra.
\newblock Joint unsupervised learning of deep representations and image
  clusters.
\newblock {\em arXiv preprint arXiv:1604.03628}, 2016.

\bibitem{Yu2015Construction}
F.~Yu, Y.~Zhang, S.~Song, A.~Seff, and J.~Xiao.
\newblock Lsun: Construction of a large-scale image dataset using deep learning
  with humans in the loop.
\newblock {\em arXiv:1506.03365}, 2015.

\bibitem{Zhou2014Place}
B.~Zhou, A.~Lapedriza, J.~Xiao, A.~Torralba, and A.~Oliva.
\newblock Learning deep features for scene recognition using places database.
\newblock In {\em NIPS}, pages 487--495, 2014.

\end{thebibliography}
}
\newpage

\appendix

\section{Supplementary Materials}
\subsection{LDPO Framework for Scene Recognition}\label{sec-method}
Here, our method is employed in the scenario of scene recognition.  As illustrated in Fig.~\ref{fig:flowchart:png}, the iteration begins by extracting the deep CNN image feature using generic ImageNet \cite{krizhevsky2012imagenet} CNN model. Next, we employed association rule mining inside the sets of either randomly grouped images (for the first iteration) or image clusters computed by ``clustering on extracted CNN features''.The top 50 mined patterns (which cover the maximum numbers of patches) per image cluster are further merged across the entire dataset to form a consolidated vocabulary of visual elements. Then, the clustering on patch-mining based feature with $k$-means is exploited. By evaluating the purity and mutual information between formed clusters in consecutive rounds, the system either terminates the current iteration (which leads to converged clustering outputs); or takes the newly refined image cluster labels to train or fine-tune the CNN model in the next iteration. 
\begin{figure*}[!]
	\begin{center}
		\includegraphics[width=1.0\linewidth]{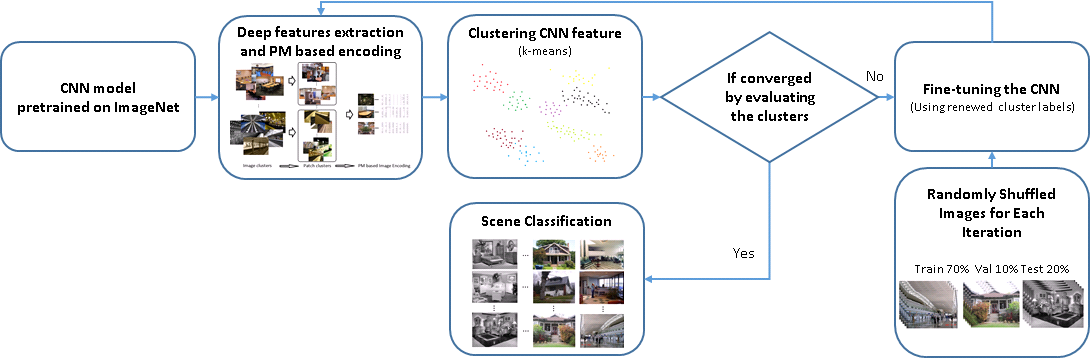}
	\end{center}
	\caption{The overview of looped deep pseudo-task optimization (LDPO) framework for scene recognition.}
	\label{fig:flowchart:png} \vspace{2mm}
\end{figure*}     

\subsection{More Experimental Results}\label{sec-Exp}\vspace{2mm}

\begin{figure}[b]
	\begin{center}
		\includegraphics[width=1\linewidth]{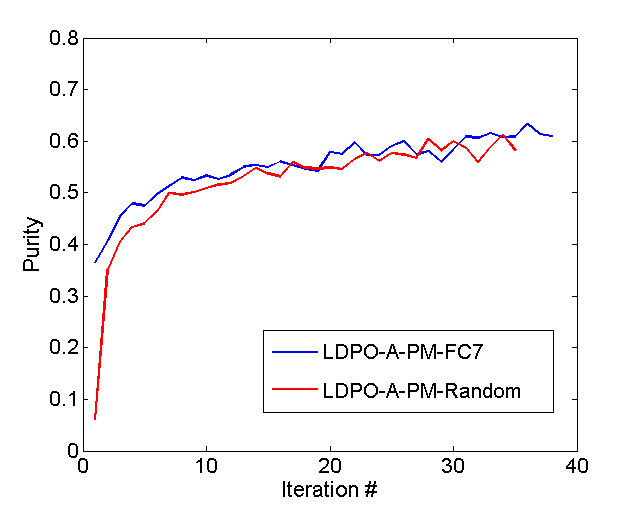}		
	\end{center}
	\caption{Clustering accuracy plots between randomly initialized clusters and ImageNet feature initialized clusters.} 
	\label{fig:cluster-init}
\end{figure}

\begin{figure*}[!t]
	\begin{center}
		\includegraphics[width=0.165\linewidth]{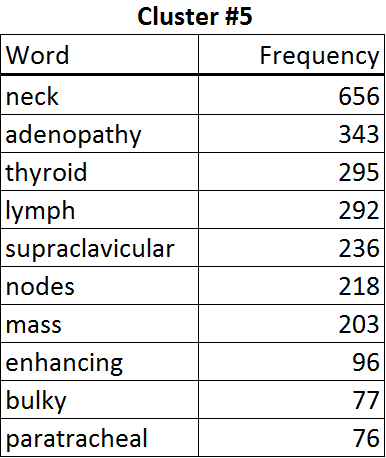}
		\includegraphics[width=0.20\linewidth]{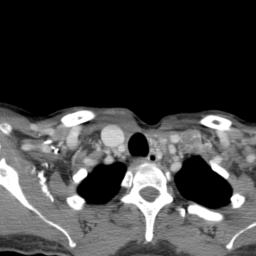}
		\includegraphics[width=0.20\linewidth]{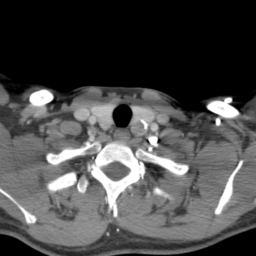}
		\includegraphics[width=0.20\linewidth]{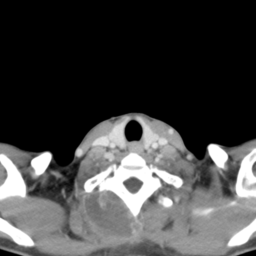}
		\includegraphics[width=0.20\linewidth]{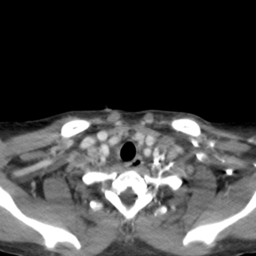}
		\includegraphics[width=0.165\linewidth]{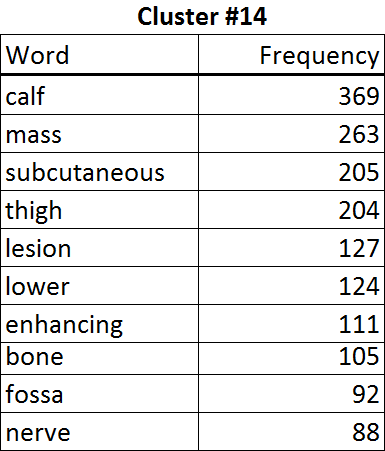}
		\includegraphics[width=0.20\linewidth]{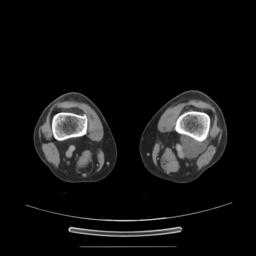}
		\includegraphics[width=0.20\linewidth]{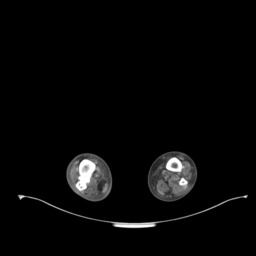}
		\includegraphics[width=0.20\linewidth]{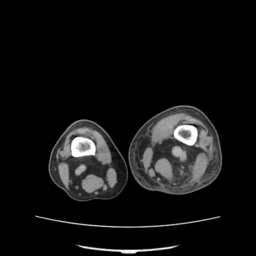}
		\includegraphics[width=0.20\linewidth]{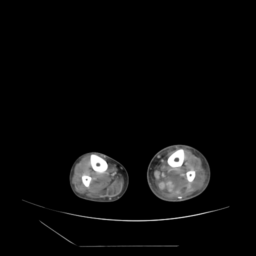}	
		\includegraphics[width=0.165\linewidth]{Cluster_ex3_t.png}
		\includegraphics[width=0.20\linewidth]{cluster_ex3_i1.png}
		\includegraphics[width=0.20 \linewidth]{cluster_ex3_i2.png}
		\includegraphics[width=0.20 \linewidth]{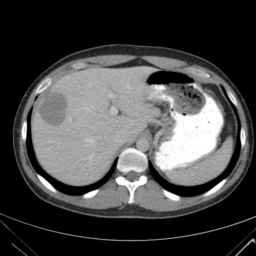}
		\includegraphics[width=0.20 \linewidth]{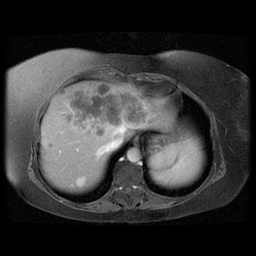}		
		\includegraphics[width=0.165\linewidth]{Cluster_ex4_t.png}
		\includegraphics[width=0.20 \linewidth]{cluster_ex4_i1.png}
		\includegraphics[width=0.20 \linewidth]{cluster_ex4_i2.png}
		\includegraphics[width=0.20 \linewidth]{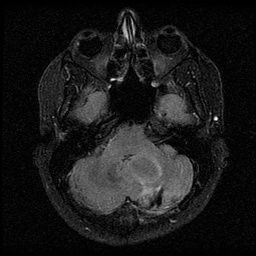}
		\includegraphics[width=0.20 \linewidth]{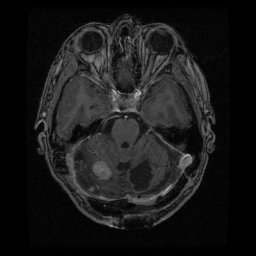}
		\includegraphics[width=0.165\linewidth]{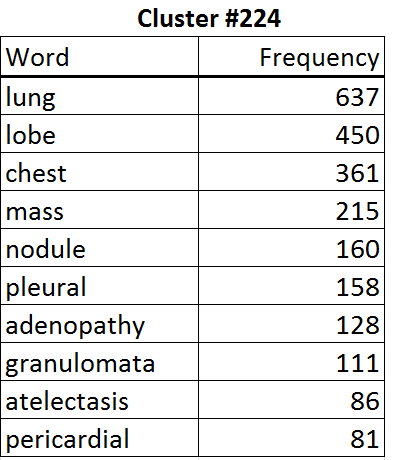}
		\includegraphics[width=0.20\linewidth]{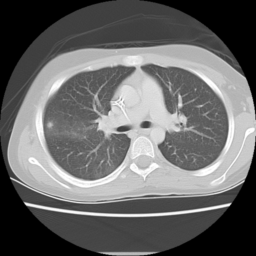}
		\includegraphics[width=0.20\linewidth]{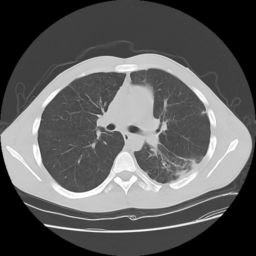}
		\includegraphics[width=0.20\linewidth]{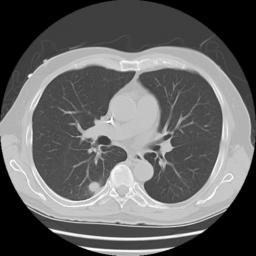}
		\includegraphics[width=0.20\linewidth]{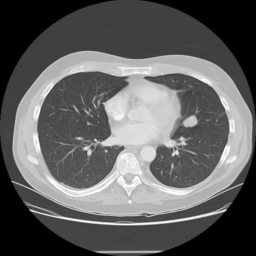}		
	\end{center}   
	\caption{Sample images of five image clusters (discovered in an unsupervised manner) with associated clinically semantic key words, containing (likely appeared) anatomies, pathologies , their attributes and imaging protocols or properties.}  
	\label{fig:Cluster_Sample:png}
\end{figure*}

\subsubsection{Unsupervised Medical Image Categorization}\label{sec-label-results} 
\begin{figure*}
	\begin{center}
		\includegraphics[width=0.32\linewidth]{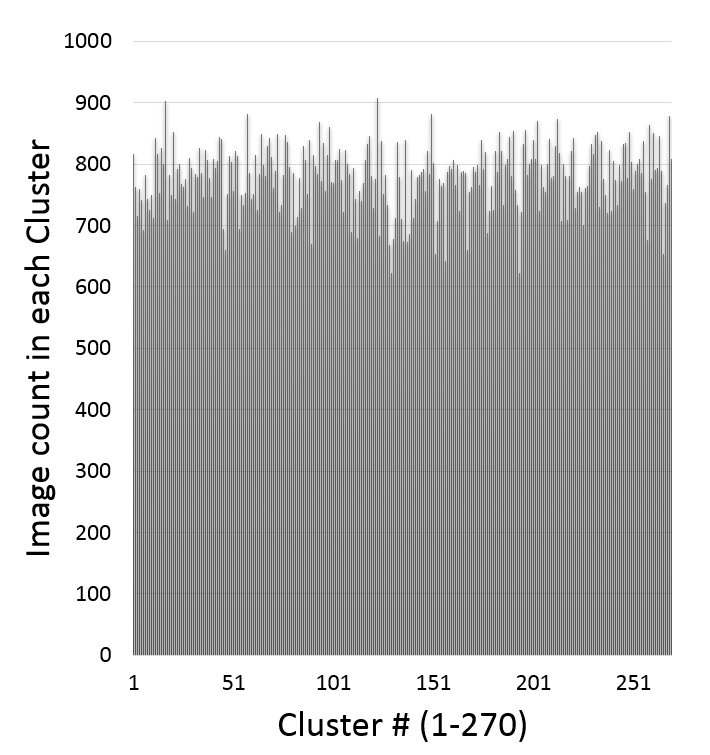}
		\includegraphics[width=0.65\linewidth]{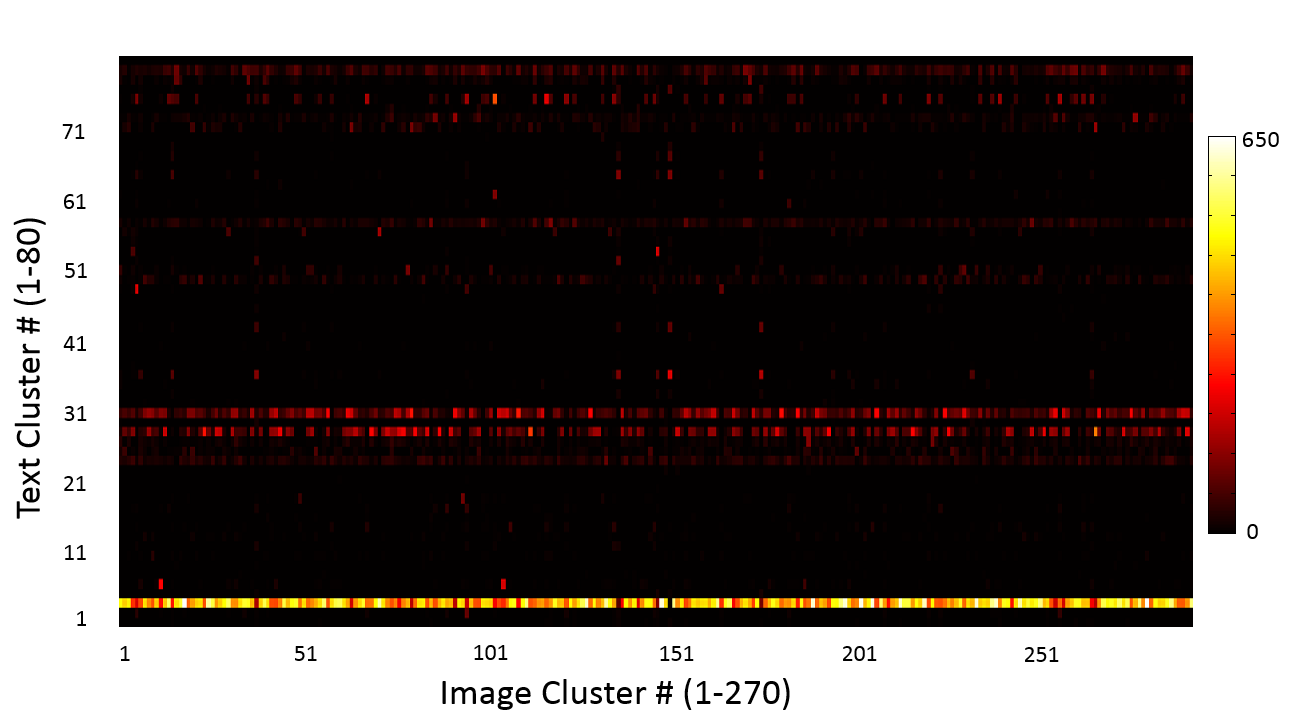}		
	\end{center}
	\caption{Affinity matrix of two clustering results (AlexNet-FC7-270 vs Text-Topics-80 produced using the approach in~\cite{Shin2015}).} 
	\label{fig:Clusters:png}
\end{figure*}

The category discovery clusters employing our LDPO method are found to be more visually coherent and cluster-wise balanced in comparison to the results in \cite{Shin2015} where clusters are formed only from text information ($\sim780K$ radiology reports). Fig.~\ref{fig:Clusters:png} {\bf Left} shows the image numbers for each cluster from the AlexNet-FC7-Topic setting. The numbers are  uniformly distributed with a mean of 778 and standard deviation of 52. Fig.~\ref{fig:Clusters:png} {\bf Right} illustrates the relation of clustering results derived from image cues or text reports~\cite{Shin2015}. Note that there is no instance-balance-per-cluster constraints in the LDPO clustering. The clusters in \cite{Shin2015} are highly uneven: 3 clusters inhabit the majority of images. Fig.~\ref{fig:Cluster_Sample:png} shows sample images and top-10 associated key words from 5 randomly selected clusters (more results in the supplementary material). The LDPO clusters are found to be semantically or clinically related to the corresponding key words, containing the information of (likely appeared) anatomies, pathologies (e.g., adenopathy, mass), their attributes (e.g., bulky, frontal) and imaging protocols or properties. 


In addition to the five sample clusters shown in the main manuscript, sample images and associated keyword labels from 20 more clusters are demonstrated and appended by the end, together with radiologist’s evaluations on each cluster in term of the subject and consistency of images.

For the space limit, we only show the results for the first 20 clusters (listed in Table 3). We hope to build a large scale publicly available database and website, similar to Micorosoft COCO, for radiology image collections: each image with its associated attritutes and labels on the clinical findings/annotatations, and with even one or two caption-like describing sentenses (extracted from original text radiology reports on RIS by advanced natural language processing techniques).
\vspace{2mm}

\subsubsection{Scene Recognition}\label{sec-scene-results}
In this section, we extend the quantitative validation of the proposed LDPO-PM method (with patch mining) on the following aspects: 1). supervised classification (e.g., Liblinear ) on image representations learned in an unsupervised fashion; 2). the convergence analysis with different initialization strategies. 

First, we evaluate the supervised discriminative power of LDPO-PM learned image representation. The MIT indoor scene dataset and its standard partition \cite{quattoni2009recognizing}, i.e., 80 training and 20 testing images per class, are adopted to examine the classification accuracy. Liblinear classification toolbox \cite{fan2008liblinear} is used on {\bf LDPO-A-PM} and {\bf LDPO-V-PM} image representation (noted as {\bf LDPO-A-PM-LL} and {\bf LDPO-V-PM-LL}) under 5-fold cross validation following \cite{Singh2012DiscPat,sun2013learning,doersch2013mid,LiLSH15CVPR,Wu2015Harvesting}. The supervised and unsupervised scene recognition accuracy results from previous state-of-the-art work and variants of our method are listed in Table  \ref{tab:classification}. The one-versus-all Liblinear classification in {\bf LDPO-A-PM-LL}, {\bf LDPO-V-PM-LL} does not noticeably improve upon purely unsupervised {\bf LDPO-A-PM} and {\bf LDPO-V-PM}. This may indicate LDPO-PM image representation are already sufficient good on separating images from different scene classes.

Next, we experiment the clustering convergence issue with two different initializations: random initialization or image labels obtained from k-means clustering on FC7 features of an ImageNet pretrained AlexNet. The clustering accuracies for both settings are plotted across iterations. As illustrated in Fig.\ref{fig:cluster-init}, the clustering accuracies of the random initialization setting boost significantly during the first several LDPO-PM iterations and finally the performances of both strategies converge to a similar level. Therefore it is evident that the LDPO convergence is insensitive to different initialization settings.
{\small
	\begin{table}[t]
		\caption{Scene Recognition Accuracy on MIT-67 Dataset \cite{quattoni2009recognizing}}
		\centering
		\begin{tabular}{|l||c|}
			\hline\hline
			{\bf Method} & {\bf Accuracy (\%)} \\[0.3ex]
			\hline\hline
			{\bf D-patch \cite{Singh2012DiscPat} }    & 38.1  \\[0.3ex]
			\hline
			{\bf D-parts \cite{sun2013learning} }    & 51.4  \\[0.3ex]
			\hline
			{\bf DMS \cite{doersch2013mid} }   & 64.0 \\[0.3ex]
			\hline
			{\bf MDPM-Alex \cite{LiLSH15CVPR}}   & 64.1 \\[0.3ex]
			\hline
			{\bf MDPM-VGG \cite{LiLSH15CVPR} }    & 77.0 \\[0.3ex]
			\hline
			{\bf MetaObject \cite{Wu2015Harvesting} }   & 78.9 \\[0.3ex]
			\hline 
			\hline
			{\bf FC (CaffeRef) }    & 57.7 \\[0.3ex]
			\hline
			{\bf FC (VGG) }   & 68.9  \\[0.3ex]
			\hline 
			{\bf CONV-FV (CaffeRef) \cite{Cimpoi2015Deep} }    & 69.7 \\[0.3ex]
			\hline
			{\bf CONV-FV (VGG) \cite{Cimpoi2015Deep}}   & {\bf 81.0} \\[0.3ex]
			\hline 
			\hline
			{\bf LDPO-A-PM-LL}    & 63.7 \\[0.3ex] 
			\hline
			{\bf LDPO-V-PM-LL}   & 72.5  \\[0.3ex]
			\hline
			{\bf LDPO-A-PM}    & 63.2  \\[0.3ex] 
			\hline
			{\bf LDPO-V-PM}   & {\bf 75.3}  \\[0.3ex]
			\hline 			
		\end{tabular}\label{tab:classification}
	\end{table}
}


\subsubsection{Computational Cost: }
LDPO runs on a node of Linux computer cluster with 16 CPU cores (x2650), 128G memory and two Nvidia K20 GPUs. The Computational costs of different method configurations (ranging from 14:35 to 28:38 in hours:minutes) are shown in Table~\ref{tab:CNN-Time} per looped iteration. The more sophisticated and feature rich settings, e.g., {\bf AlexNet-Conv5-FV}, {\bf AlexNet-Conv5-VLAD} and {\bf VGG-VD-FC7-PM}, require more time to converge. 

{\small\small
\begin{table}
	\caption{Computational Cost of LDPO}
	\centering
	\begin{tabular}{|l||c|}
		\hline\hline
		{\bf CNN setting} & {\bf Time per iter.(HH:MM)} \\[0.3ex] 
		\hline\hline
		\multicolumn{2}{|c|}{Medical Image Categorization}\\
		\hline
		{\bf AlexNet-FC7-Topic}    & 14:35  \\[0.3ex]
		\hline
		{\bf AlexNet-FC7-Imagenet}   & 14:40  \\[0.3ex]
		\hline
		{\bf AlexNet-Conv5-FV}   & 17:40  \\[0.3ex]
		\hline
		{\bf AlexNet-Conv5-VLAD}   & 15:44  \\[0.3ex]
		\hline
		{\bf GoogLeNet-Pool5}    & 21:12  \\[0.3ex]
		\hline
		{\bf GoogLeNet-Inc.5b-VLAD}   & 23:35  \\[0.3ex]
		\hline\hline
		\multicolumn{2}{|c|}{Scene Recognition}\\
		\hline
		{\bf AlexNet-FC7-PM}    & 18:11  \\[0.3ex]
		\hline
		{\bf VGG-VD-FC7-PM}    & 28:38  \\[0.3ex] 
		\hline
	\end{tabular}\label{tab:CNN-Time}
\end{table}}

\cleardoublepage

\hspace{1mm}   
\newpage


\section*{Supplementary material}

Table 3. *20 clusters with Radiologist's Evaluation*

| Cluster Number | Radiologist's Impression (Label) | Withih-Class Coherence or Consistency |
|---|---|---|
| 1 | Abdomen or pelvis CT with disease | High |
| 2 | Abdomen or pelvis CT with disease | Moderately High |
| 3 | Abdomen or pelvis CT with disease (predominantly bowel) | Moderately High |
| 4 | Chest CT with pulmonary disease | High |
| 5 | Neck or chest CT with tumor or lymphadenopathy | High |
| 6 | Abdomen or pelvis CT with disease | High |
| 7 | Chest CT containing mass | Moderately High |
| 8 | Chest CT with pulmonary disease (predominantly pleural) | Moderately High |
| 9 | Body MRI (chest, abdomen, or pelvis) with mass/lesion | High |
| 10 | Chest or abdomen CT (chest, abdomen, or pelvis) with lesion in hepatic dome | High |
| 11 | Chest CT with lesions (predominately pleural) | High |
| 12 | Abdomen CT with lesions (predominately hepatic or renal) | High |
| 13 | Abdominal or Pelvis MRI with tumor | High |
| 14 | Lower extremity CT with disease | Moderately Low (2 internal sub-classes) |
| 15 | Chest CT with nodule/mass | High |
| 16 | Abdominal MRI with tumor (predominately hepatic) | Moderately High |
| 17 | Abdomen CT with hepatic lesions | High |
| 18 | Pelvis CT with lesions/tumor | High |
| 19 | Chest CT with pulmonary disease (predominately pleural) | High |
| 20 | Chest CT with tumor | High |

# Cluster #001

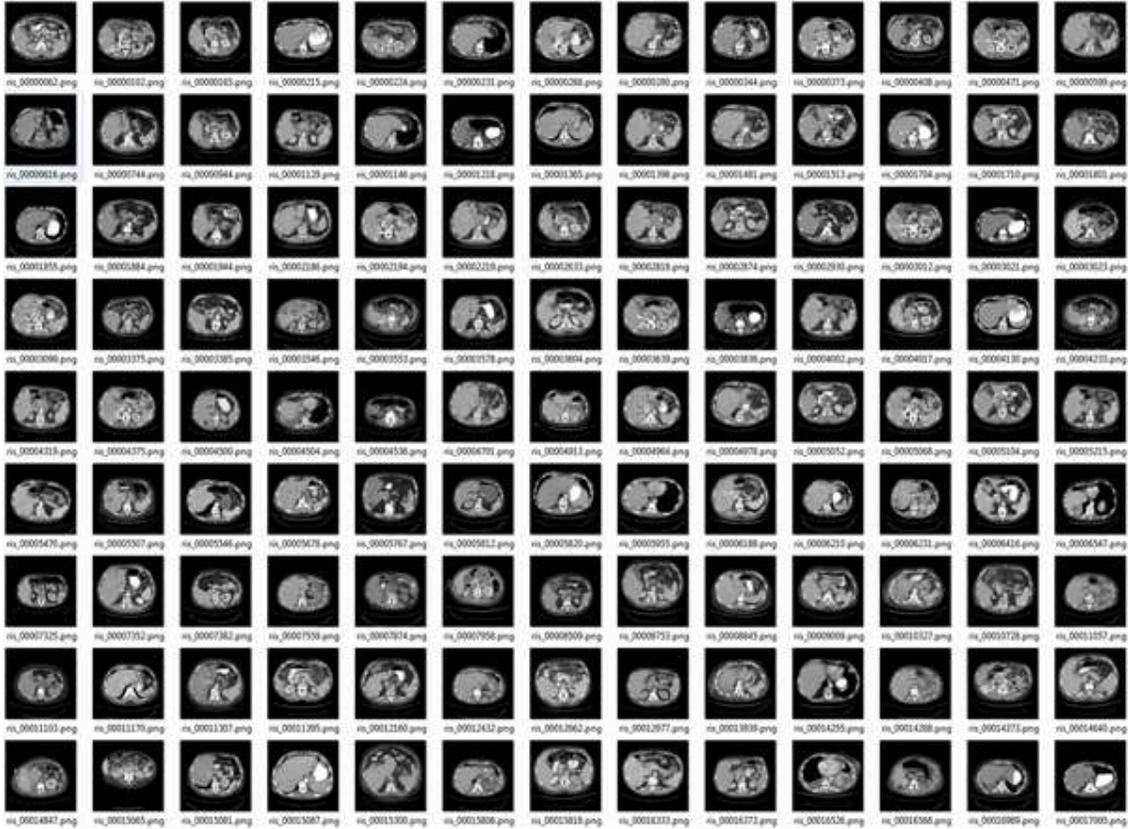

| WORD | Frequency | Relative Frequency |
|---|---|---|
| abdomen | 308 | 0.84% |
| unchanged | 306 | 0.84% |
| liver | 294 | 0.80% |
| mass | 253 | 0.69% |
| pelvis | 193 | 0.53% |
| since | 190 | 0.52% |
| adrenal | 149 | 0.41% |
| evidence | 145 | 0.40% |
| pancreas | 144 | 0.39% |
| lobe | 142 | 0.39% |

Radiologist's Evaluation
Category Subject: Abdomen or pelvis CT with disease
Consistency: High

# Cluster #002

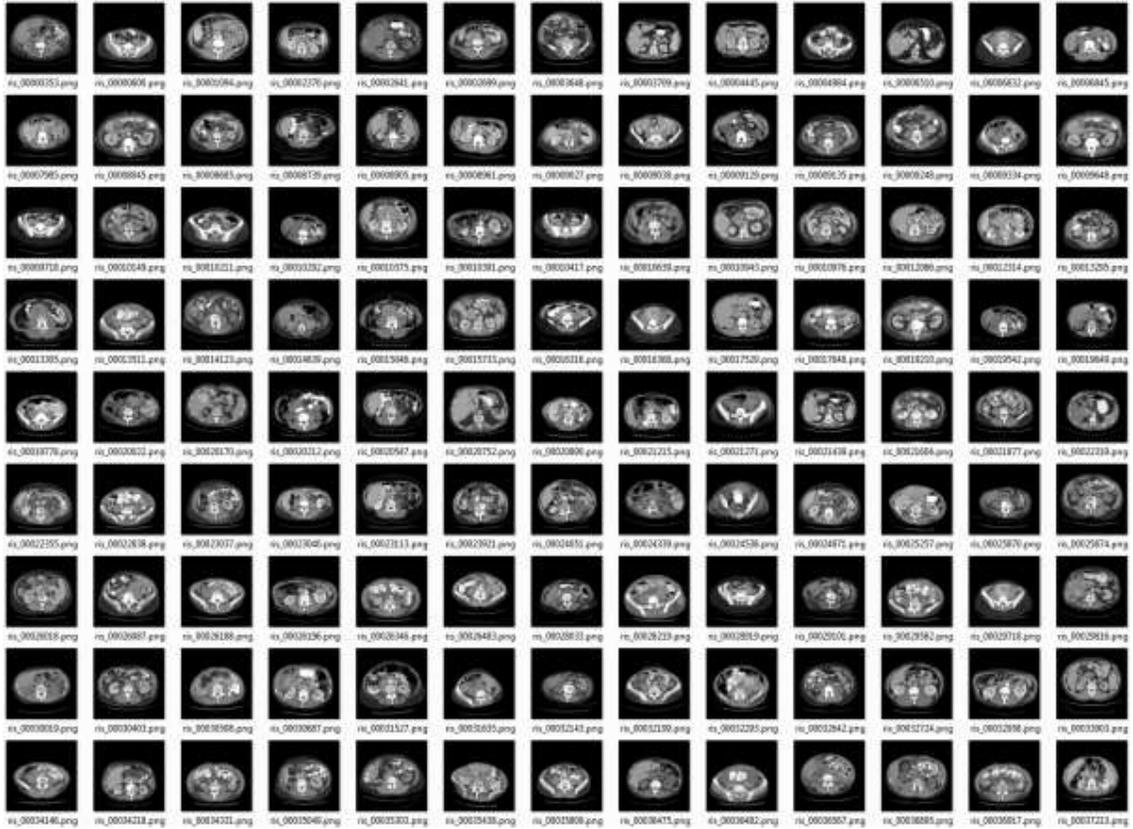

| WORD | Frequency | Relative Frequency |
|---|---|---|
| abdomen | 221 | 0.94% |
| pelvis | 199 | 0.84% |
| mass | 168 | 0.71% |
| kidney | 153 | 0.65% |
| adenopathy | 137 | 0.58% |
| liver | 127 | 0.54% |
| lymph | 86 | 0.36% |
| mesenteric | 85 | 0.36% |
| abdominal | 82 | 0.35% |
| evidence | 81 | 0.34% |

Radiologist's Evaluation
Category Subject: Abdomen or pelvis CT with disease
Consistency: Moderately High

# Cluster #003

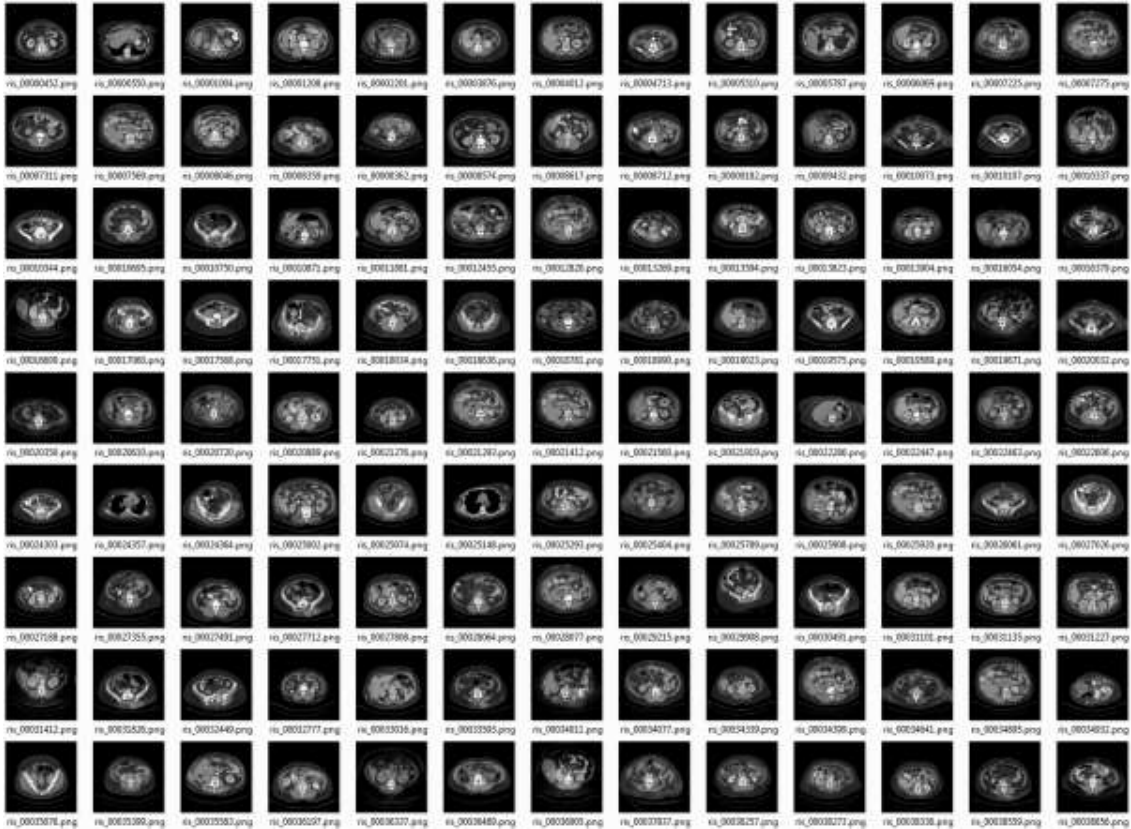

| WORD | Frequency | Relative Frequency |
|---|---|---|
| pelvis | 275 | 0.78% |
| abdomen | 249 | 0.71% |
| adenopathy | 182 | 0.52% |
| mass | 181 | 0.52% |
| liver | 166 | 0.47% |
| kidney | 165 | 0.47% |
| lesion | 129 | 0.37% |
| fluid | 124 | 0.35% |
| enhancing | 123 | 0.35% |
| bowel | 121 | 0.34% |

Radiologist's Evaluation
Category Subject: Abdomen or pelvis CT with disease (predominantly bowel)
Consistency: Moderately High

# Cluster #004

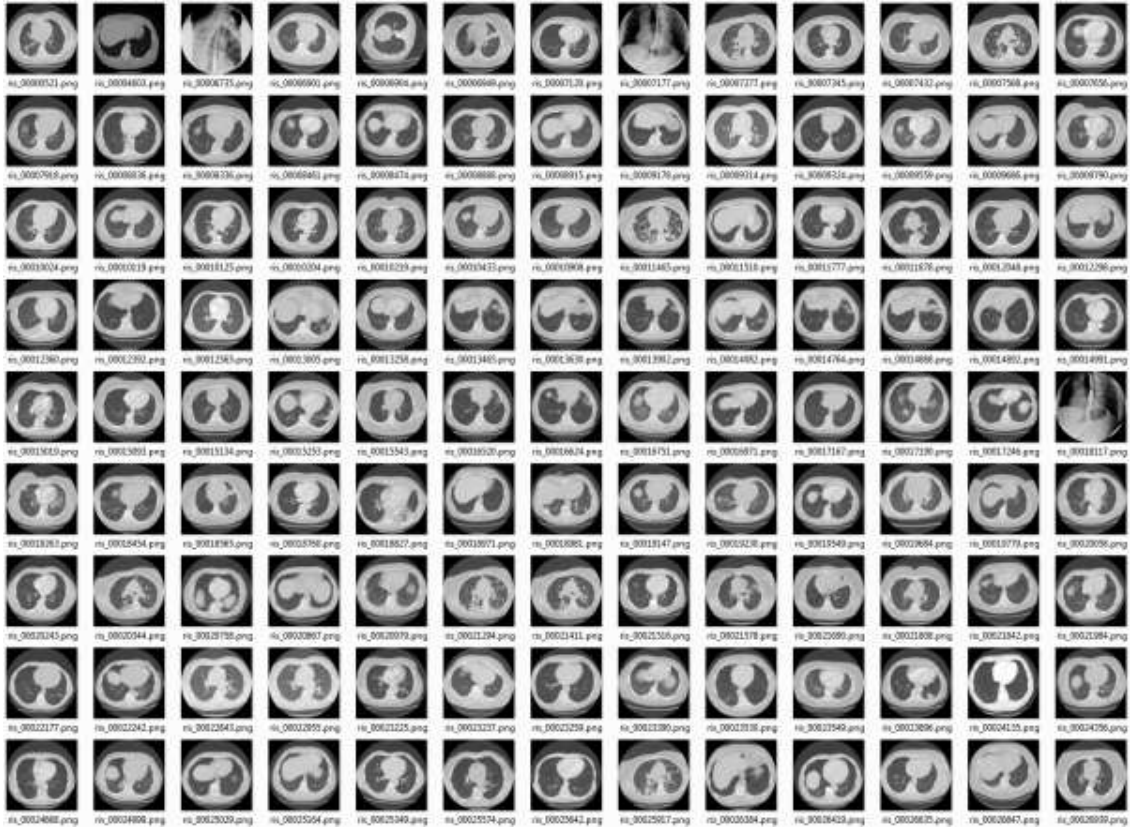

| WORD | Frequency | Relative Frequency |
|---|---|---|
| lung | 825 | 2.43% |
| lobe | 369 | 1.09% |
| chest | 288 | 0.85% |
| pleural | 207 | 0.61% |
| infiltrate | 205 | 0.60% |
| nodule | 155 | 0.46% |
| atelectasis | 140 | 0.41% |
| adenopathy | 105 | 0.31% |
| axilla | 102 | 0.30% |
| effusion | 92 | 0.27% |

Radiologist's Evaluation
Category Subject: Chest CT with pulmonary disease
Consistency: High

# Cluster #005

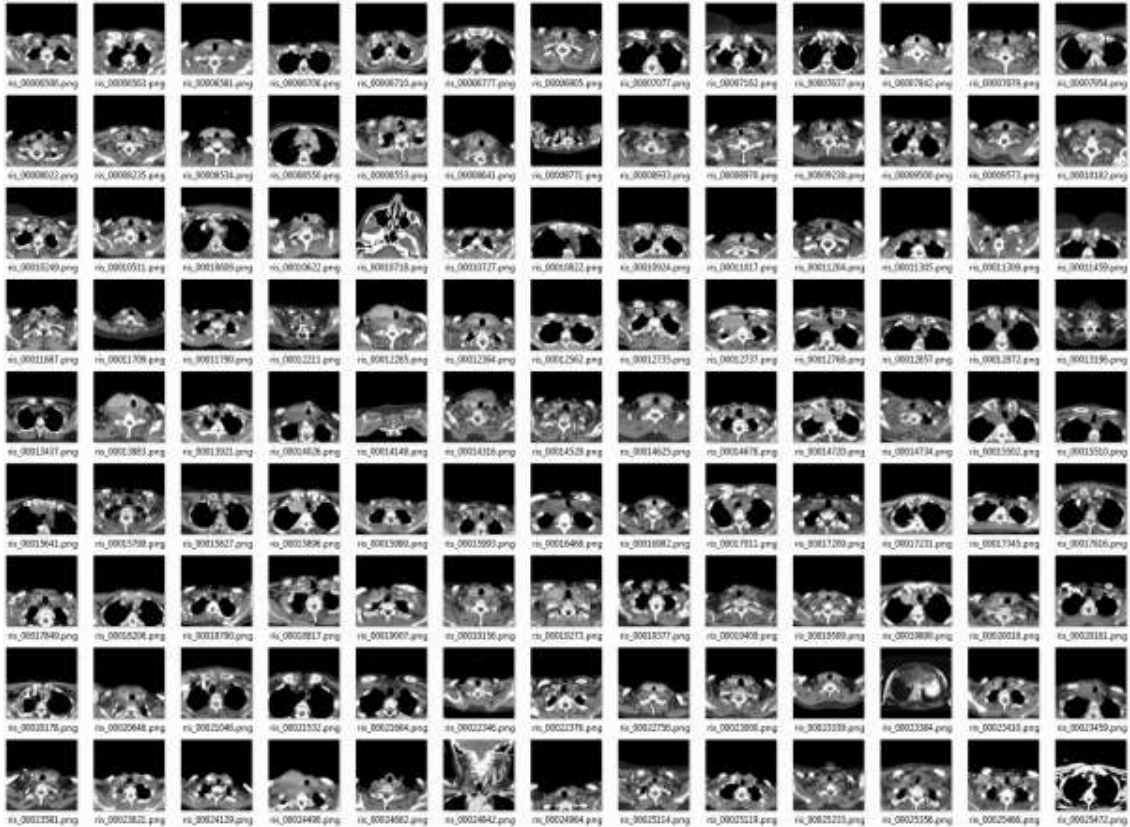

| WORD | Frequency | Relative Frequency |
|---|---|---|
| neck | 656 | 1.92% |
| adenopathy | 343 | 1.00% |
| thyroid | 295 | 0.86% |
| lymph | 292 | 0.86% |
| supraclavicular | 236 | 0.69% |
| nodes | 218 | 0.64% |
| mass | 203 | 0.59% |
| enhancing | 96 | 0.28% |
| bulky | 77 | 0.23% |
| paratracheal | 76 | 0.22% |

Radiologist's Evaluation
Category Subject: Neck or chest CT with tumor or lymphadenopathy
Consistency: High

# Cluster #006

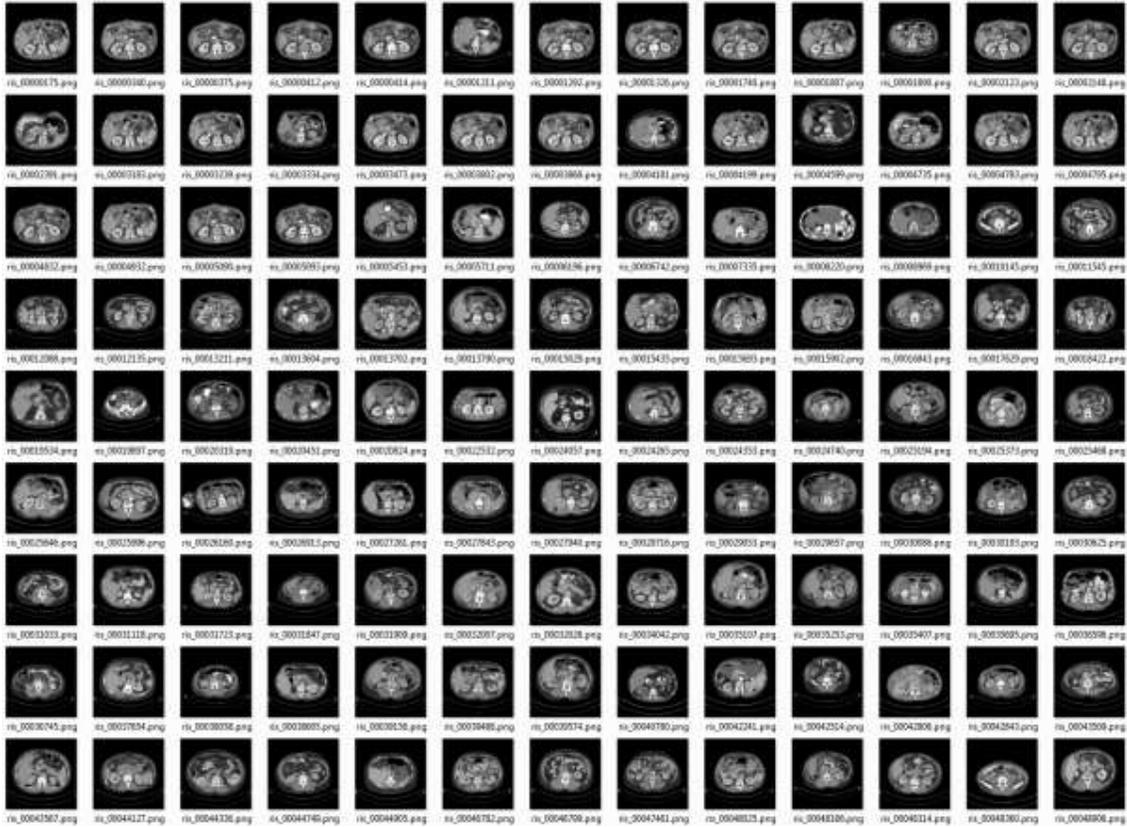

| WORD | Frequency | Relative Frequency |
|---|---|---|
| bowel | 355 | 1.04% |
| gas | 338 | 0.99% |
| pelvis | 325 | 0.95% |
| abdomen | 322 | 0.94% |
| fluid | 278 | 0.81% |
| dilation | 242 | 0.71% |
| colon | 240 | 0.70% |
| mass | 183 | 0.54% |
| liver | 177 | 0.52% |
| renal | 172 | 0.50% |

Radiologist's Evaluation
Category Subject: Abdomen or pelvis CT with disease
Consistency: High

# Cluster #007

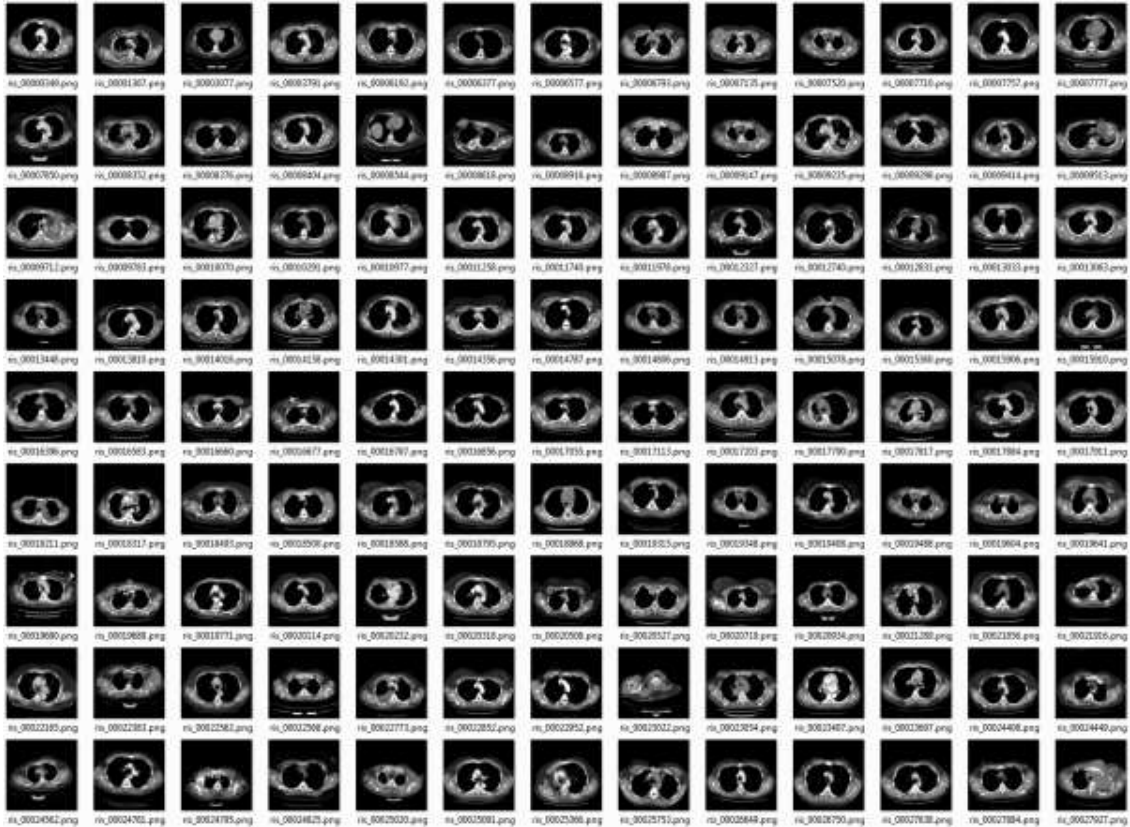

| WORD | Frequency | Relative Frequency |
|---|---|---|
| adenopathy | 406 | 1.21% |
| chest | 403 | 1.20% |
| lung | 374 | 1.11% |
| pleural | 276 | 0.82% |
| mass | 264 | 0.79% |
| axilla | 227 | 0.68% |
| lymph | 211 | 0.63% |
| effusion | 188 | 0.56% |
| lobe | 166 | 0.49% |
| mediastinum | 149 | 0.44% |

Radiologist's Evaluation
Category Subject: Chest CT containing mass
Consistency: Moderate

# Cluster #008

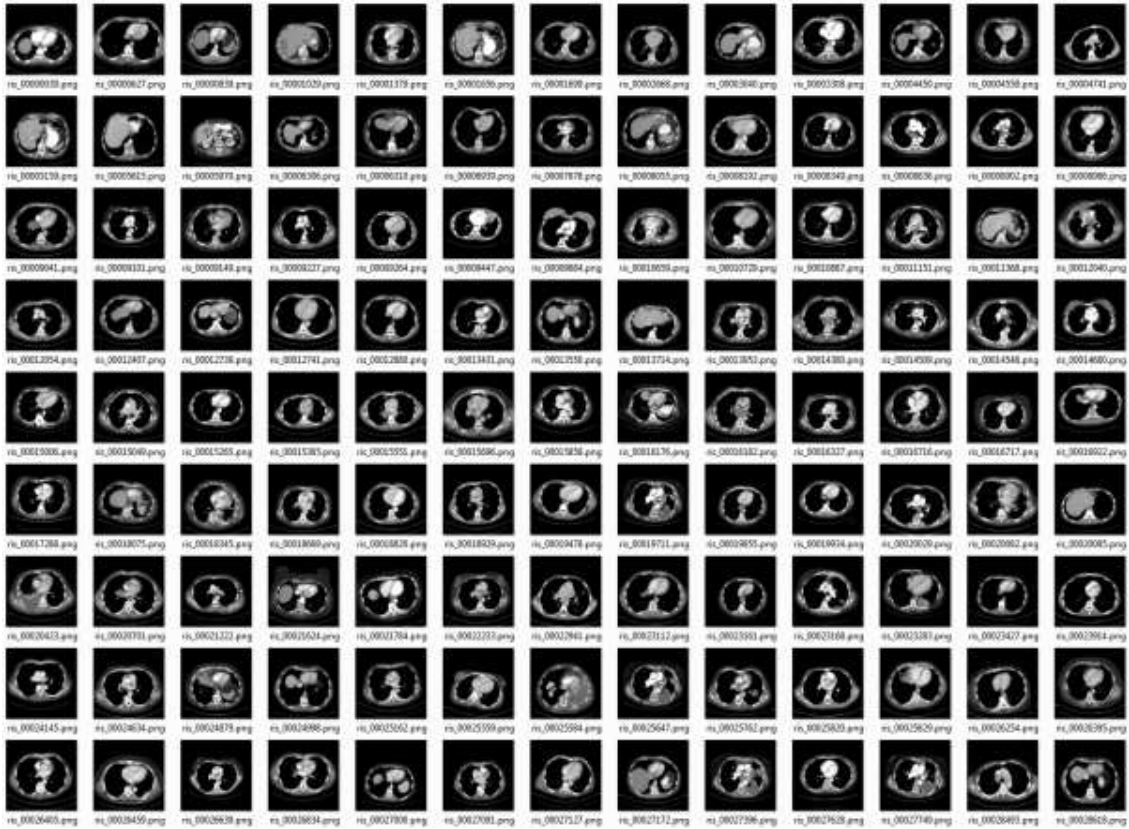

| WORD | Frequency | Relative Frequency |
|---|---|---|
| lobe | 456 | 1.22% |
| lung | 301 | 0.80% |
| mass | 282 | 0.75% |
| adenopathy | 258 | 0.69% |
| pleural | 252 | 0.67% |
| chest | 235 | 0.63% |
| pulmonary | 167 | 0.45% |
| effusion | 161 | 0.43% |
| nodule | 158 | 0.42% |
| masses | 150 | 0.40% |

Radiologist's Evaluation
Category Subject: Chest CT with pulmonary disease (predominantly pleural)
Consistency: Moderate

# Cluster #009

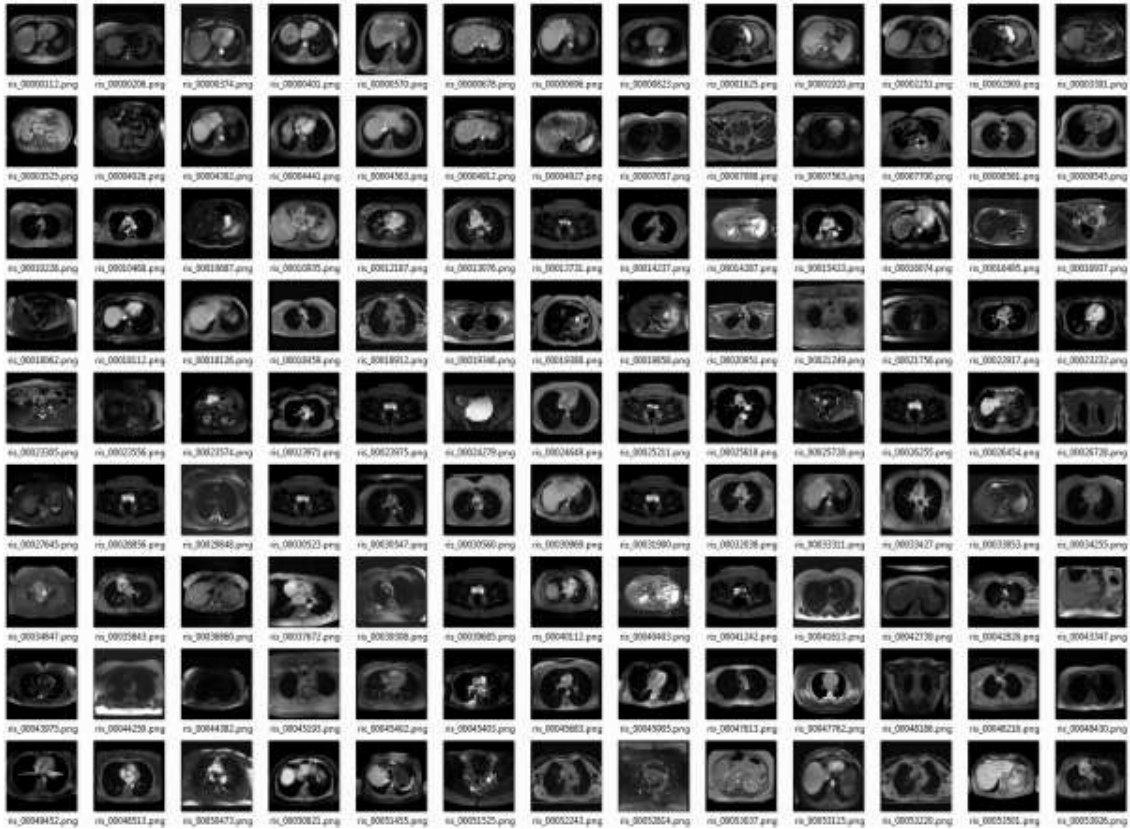

| WORD | Frequency | Relative Frequency |
|---|---|---|
| liver | 446 | 0.98% |
| mass | 270 | 0.59% |
| enhancing | 239 | 0.52% |
| focus | 235 | 0.52% |
| chest | 228 | 0.50% |
| prolonged | 221 | 0.48% |
| abdomen | 217 | 0.48% |
| enhancement | 214 | 0.47% |
| lesion | 211 | 0.46% |
| evidence | 209 | 0.46% |

Radiologist's Evaluation
Category Subject: Body MRI (chest, abdomen, or pelvis) with mass/lesion
Consistency: High

# Cluster #010

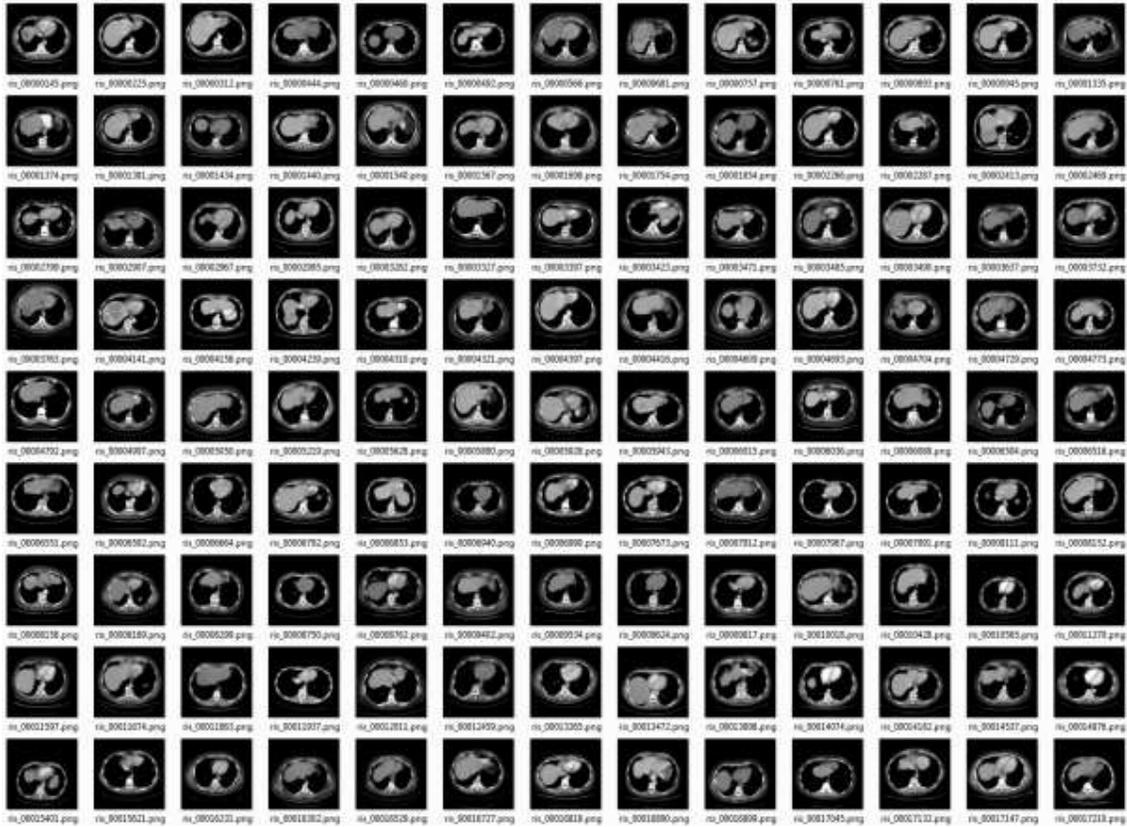

| WORD | Frequency | Relative Frequency |
|---|---|---|
| liver | 360 | 1.08% |
| lobe | 256 | 0.77% |
| lung | 238 | 0.71% |
| mass | 216 | 0.65% |
| pleural | 205 | 0.61% |
| abdomen | 159 | 0.48% |
| chest | 148 | 0.44% |
| pelvis | 147 | 0.44% |
| effusion | 142 | 0.43% |
| evidence | 138 | 0.41% |

Radiologist's Evaluation
Category Subject: Chest or abdomen CT (chest, abdomen, or pelvis) with lesion in hepatic dome
Consistency: High

# Cluster #011

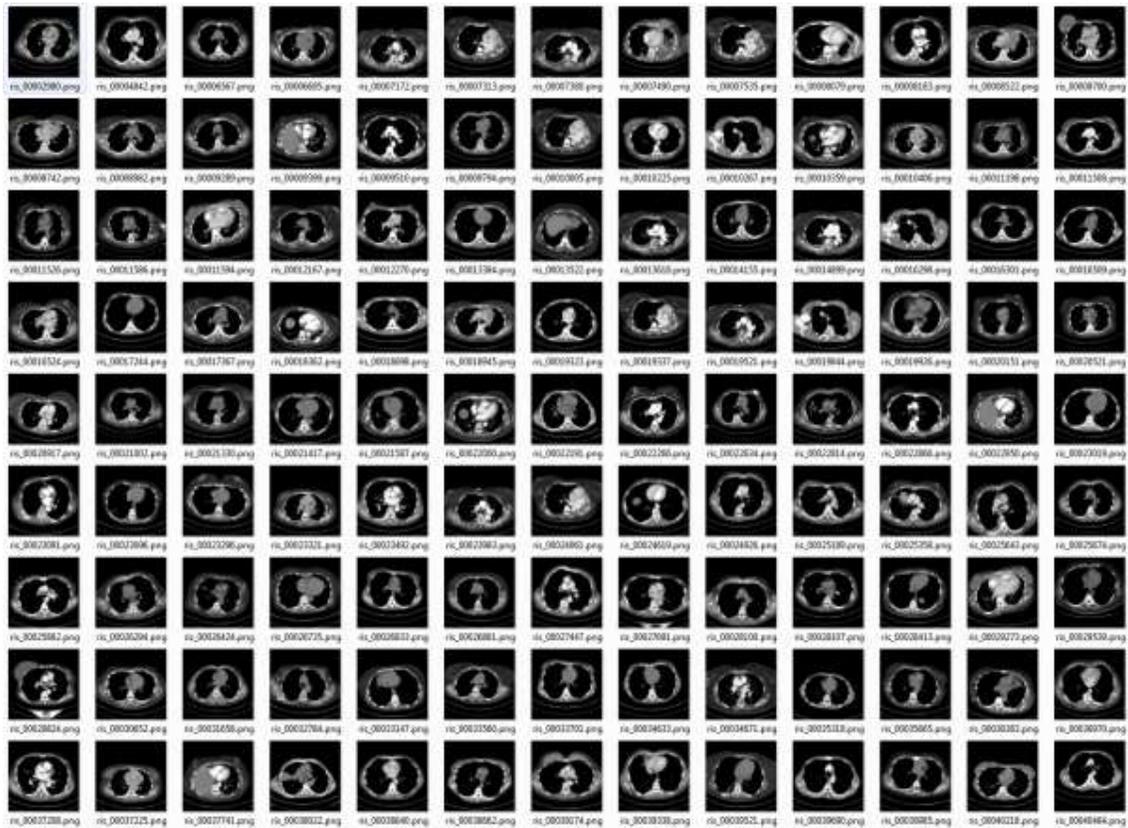

| WORD | Frequency | Relative Frequency |
|---|---|---|
| lobe | 815 | 1.93% |
| lung | 516 | 1.22% |
| nodule | 272 | 0.65% |
| pleural | 247 | 0.59% |
| pulmonary | 224 | 0.53% |
| chest | 198 | 0.47% |
| nodules | 192 | 0.46% |
| adenopathy | 175 | 0.42% |
| mass | 156 | 0.37% |
| ct | 147 | 0.35% |

Radiologist's Evaluation
Category Subject: Chest CT with lesions (predominately pleural)
Consistency: Moderately High

# Cluster #012

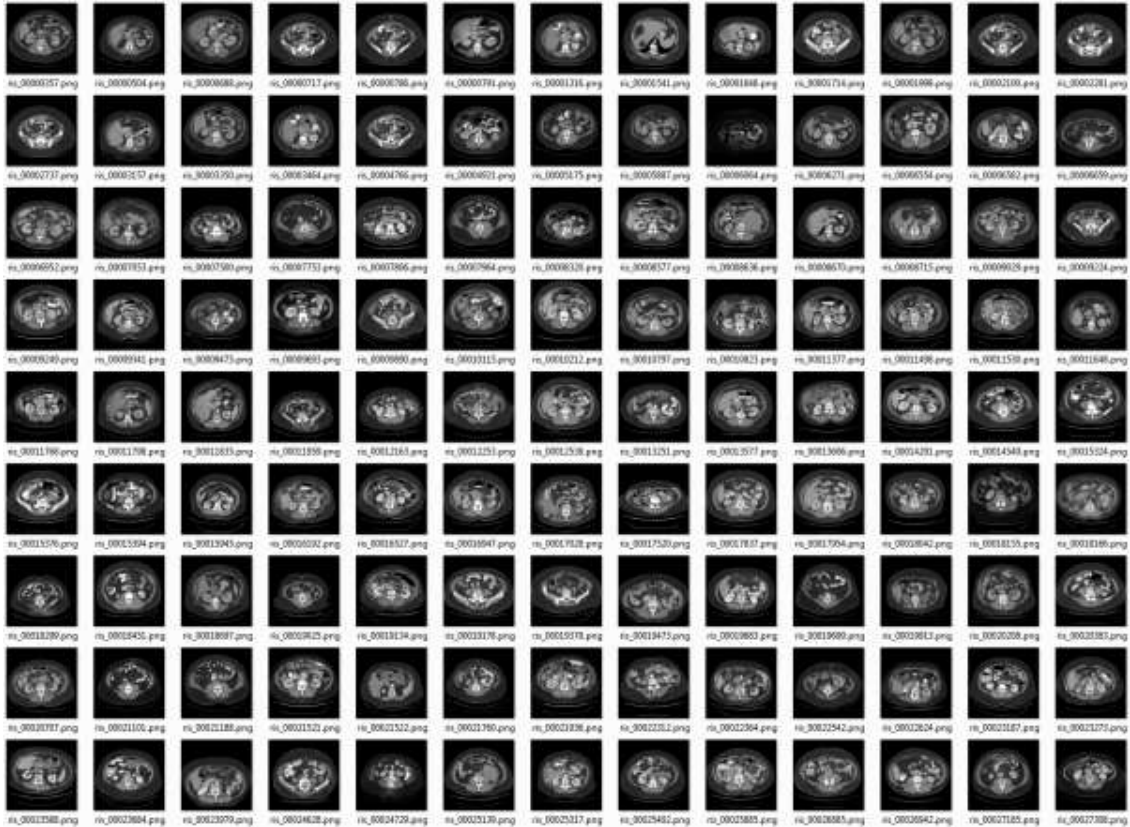

| WORD | Frequency | Relative Frequency |
|---|---|---|
| kidney | 293 | 0.86% |
| abdomen | 252 | 0.74% |
| adenopathy | 248 | 0.73% |
| mass | 241 | 0.71% |
| pelvis | 212 | 0.62% |
| lesion | 174 | 0.51% |
| enhancing | 169 | 0.50% |
| evidence | 159 | 0.47% |
| renal | 148 | 0.43% |
| foci | 130 | 0.38% |

Radiologist's Evaluation
Category Subject: Abdomen CT with lesions (predominately hepatic or renal)
Consistency: Moderately High

# Cluster #013

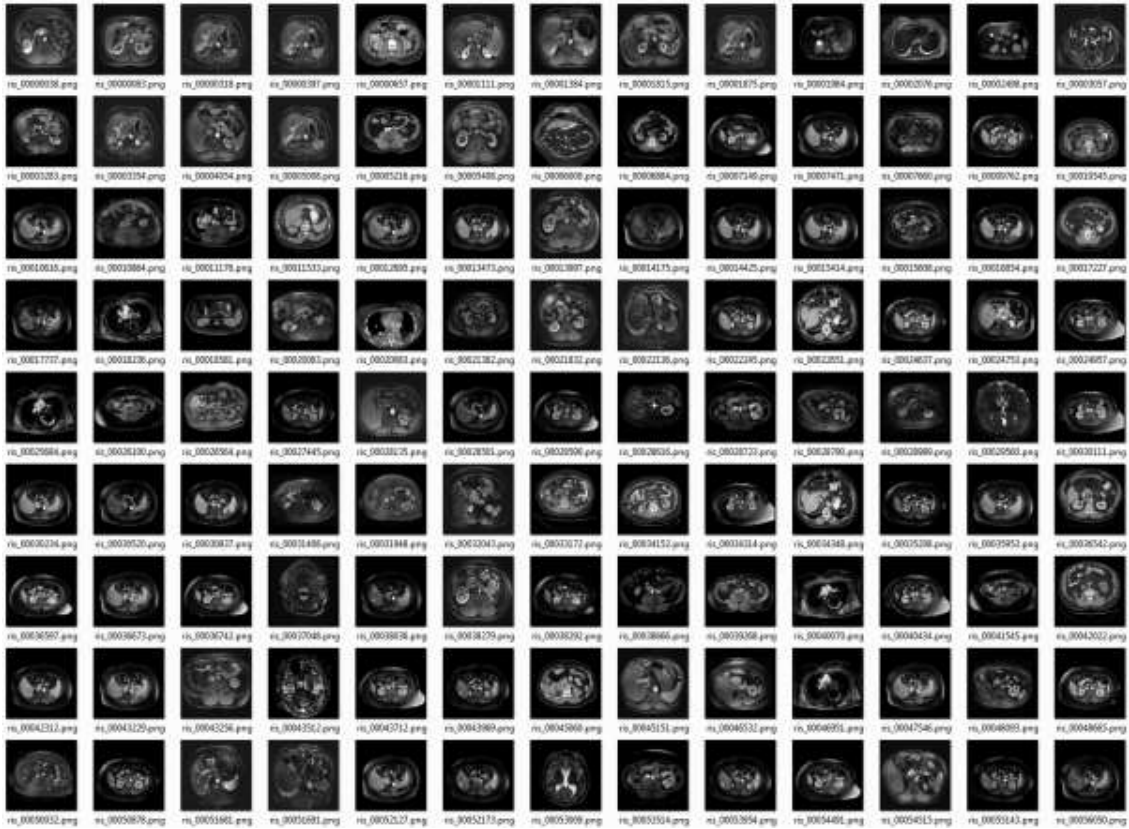

| WORD | Frequency | Relative Frequency |
|---|---|---|
| t2 | 341 | 0.81% |
| kidney | 338 | 0.81% |
| enhancing | 327 | 0.78% |
| liver | 323 | 0.77% |
| abdomen | 273 | 0.65% |
| mass | 231 | 0.55% |
| enhancement | 174 | 0.42% |
| lesion | 160 | 0.38% |
| metastases | 155 | 0.37% |
| hyperintense | 153 | 0.37% |

Radiologist's Evaluation
Category Subject: Abdominal or Pelvis MRI with tumor
Consistency: High

# Cluster #014

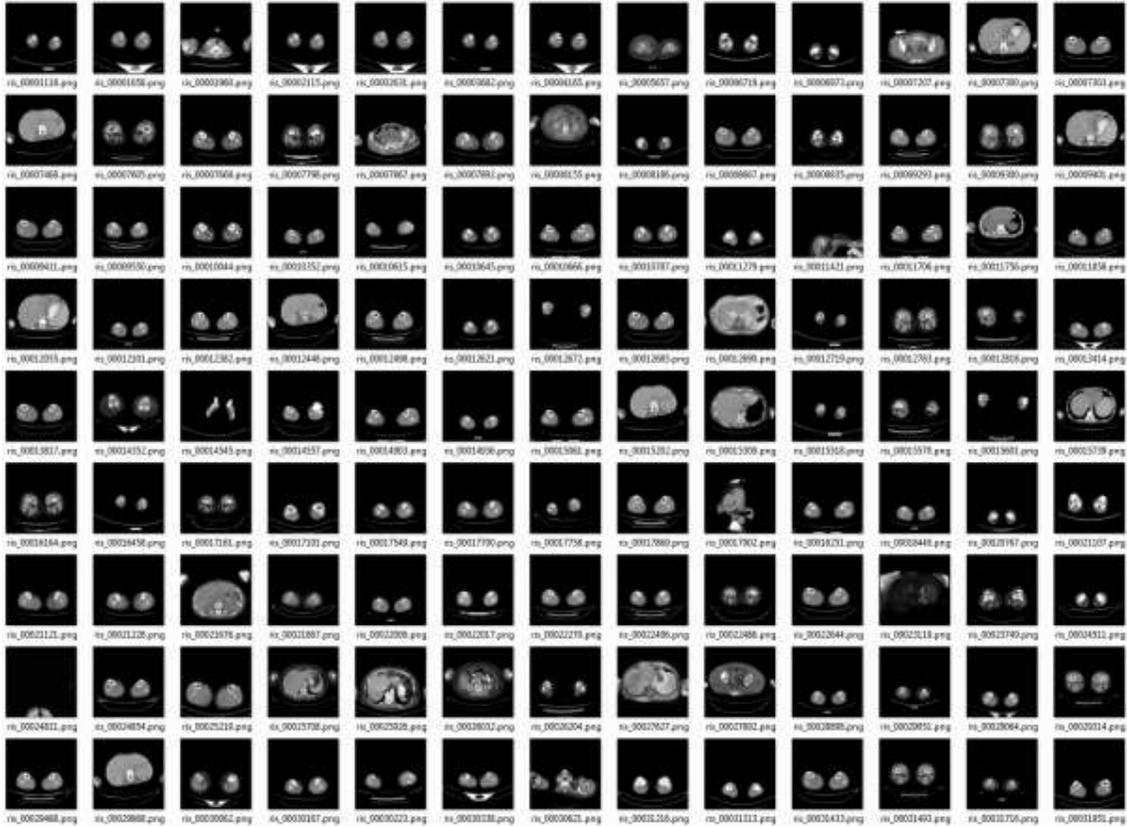

| WORD | Frequency | Relative Frequency |
|---|---|---|
| calf | 369 | 1.10% |
| mass | 263 | 0.78% |
| subcutaneous | 205 | 0.61% |
| thigh | 204 | 0.61% |
| lesion | 127 | 0.38% |
| lower | 124 | 0.37% |
| enhancing | 111 | 0.33% |
| bone | 105 | 0.31% |
| fossa | 92 | 0.27% |
| nerve | 88 | 0.26% |

Radiologist's Evaluation
Category Subject: Lower extremity CT with disease
Consistency: Moderate-Low

# Cluster #015

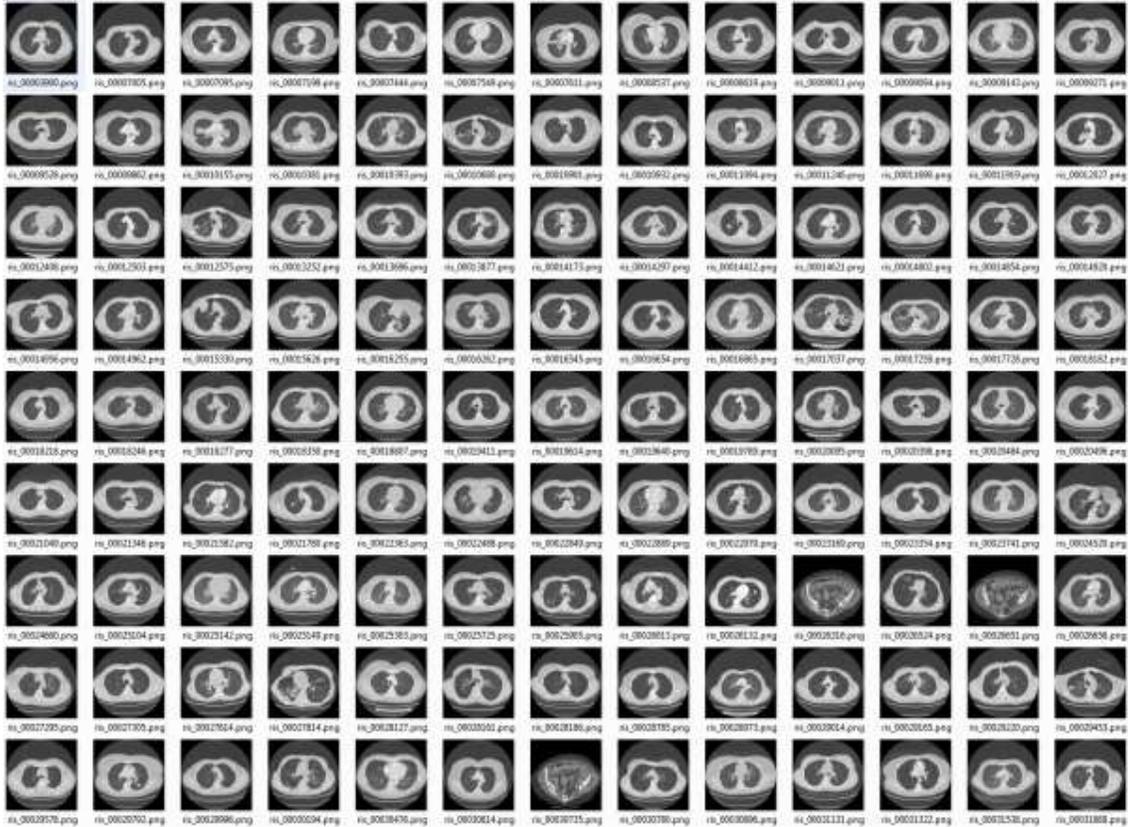

| WORD | Frequency | Relative Frequency |
|---|---|---|
| lung | 676 | 2.14% |
| lobe | 446 | 1.41% |
| chest | 299 | 0.95% |
| nodule | 245 | 0.78% |
| mass | 193 | 0.61% |
| infiltrate | 163 | 0.52% |
| pleural | 154 | 0.49% |
| scarring | 114 | 0.36% |
| adenopathy | 105 | 0.33% |
| ct | 97 | 0.31% |

Radiologist's Evaluation
Category Subject: Chest CT with nodule/mass
Consistency: High

# Cluster #016

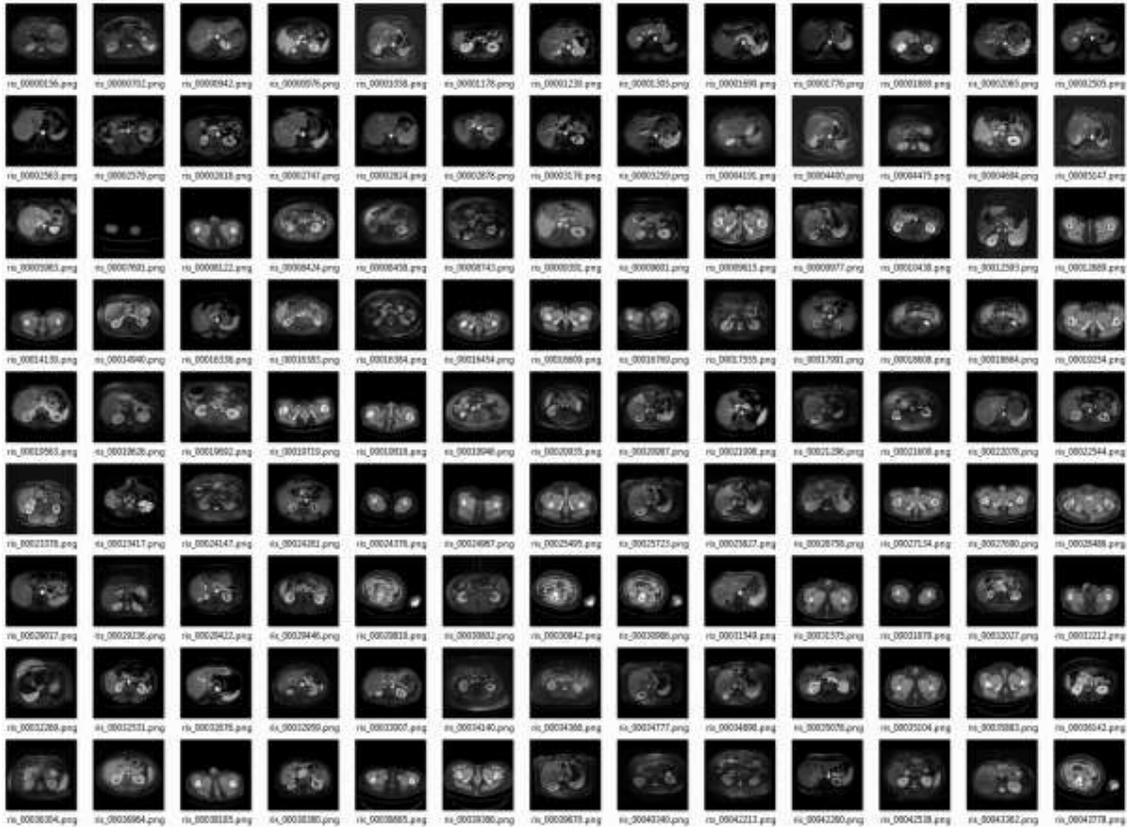

| WORD | Frequency | Relative Frequency |
|---|---|---|
| liver | 406 | 0.90% |
| kidney | 383 | 0.85% |
| mass | 326 | 0.72% |
| lesion | 318 | 0.70% |
| enhancing | 286 | 0.63% |
| lesions | 220 | 0.49% |
| enhancement | 210 | 0.46% |
| abdomen | 194 | 0.43% |
| renal | 155 | 0.34% |
| cysts | 127 | 0.28% |

Radiologist's Evaluation
Category Subject: Abdominal MRI with tumor (predominately hepatic)
Consistency: Moderately High

# Cluster #017

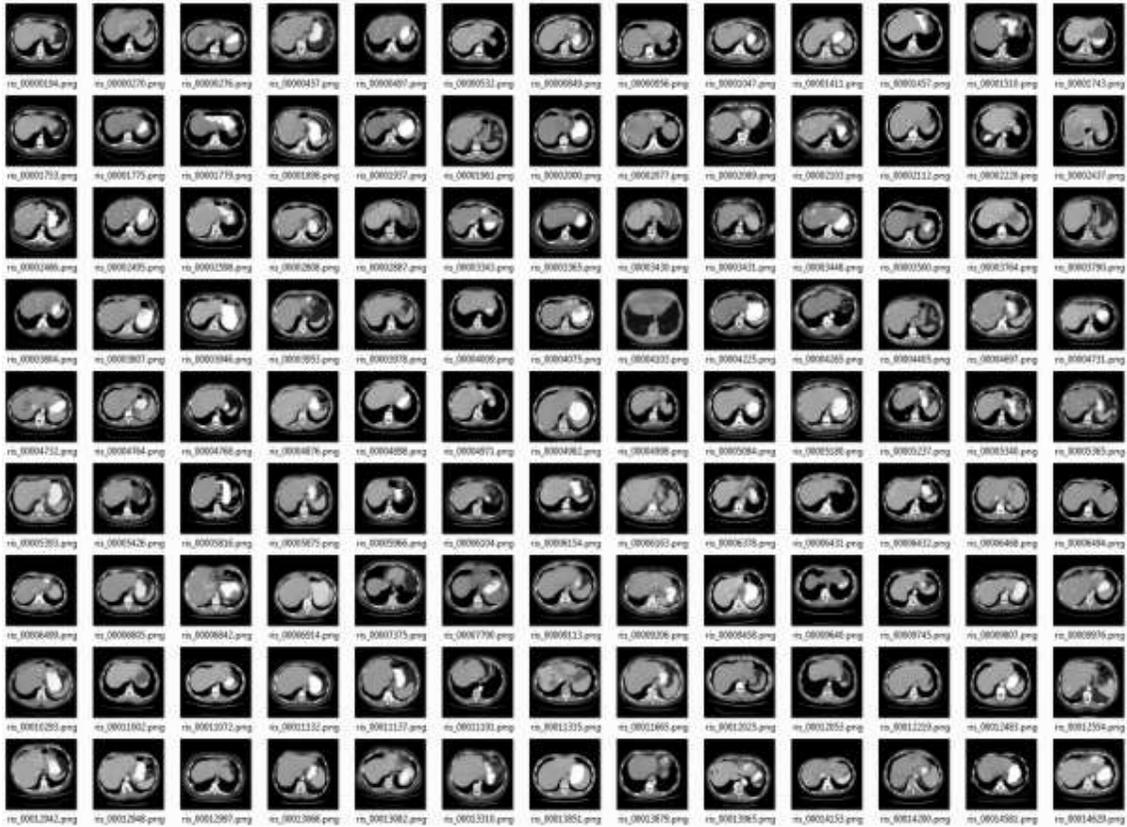

| WORD | Frequency | Relative Frequency |
|---|---|---|
| liver | 515 | 1.68% |
| foci | 253 | 0.82% |
| abdomen | 242 | 0.79% |
| pelvis | 237 | 0.77% |
| adenopathy | 191 | 0.62% |
| focus | 183 | 0.60% |
| lobe | 158 | 0.51% |
| mass | 156 | 0.51% |
| spleen | 147 | 0.48% |
| lesion | 120 | 0.39% |

Radiologist's Evaluation
Category Subject: Abdomen CT with hepatic lesions
Consistency: High

# Cluster #018

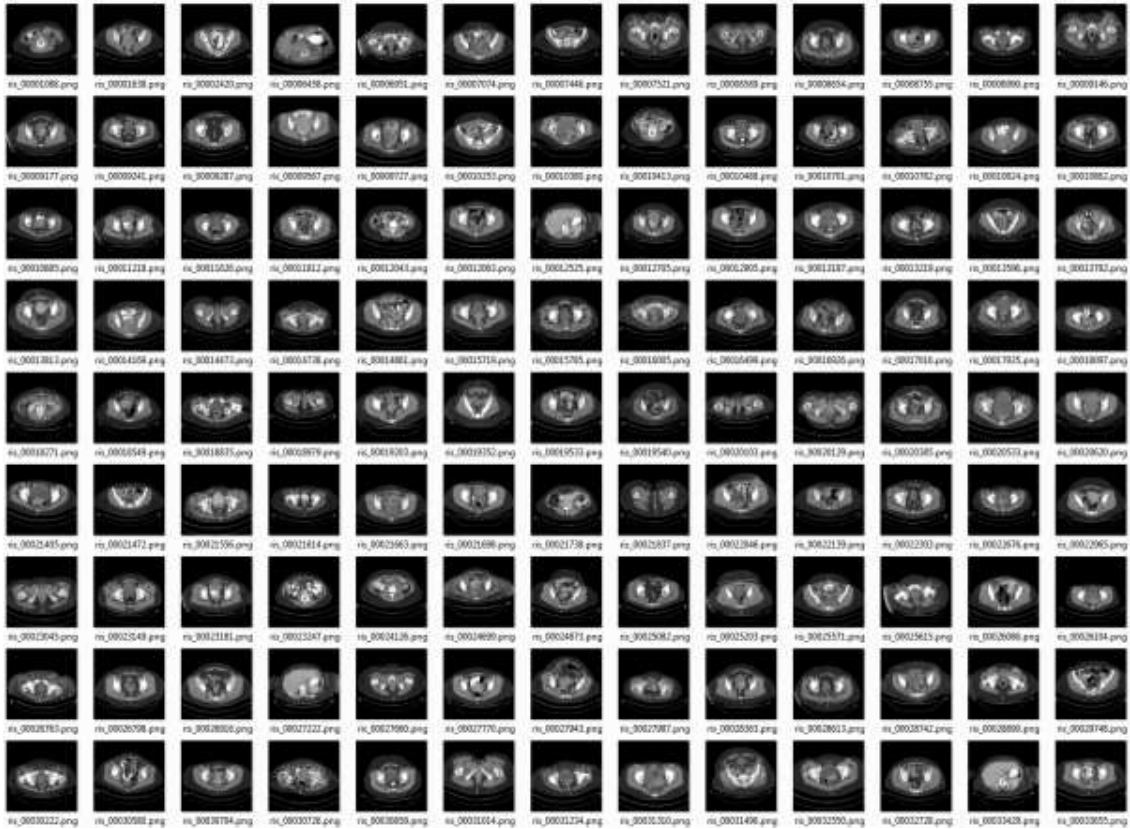

| WORD | Frequency | Relative Frequency |
|---|---|---|
| **pelvis** | 323 | 0.88% |
| **mass** | 185 | 0.50% |
| **adenopathy** | 180 | 0.49% |
| **iliac** | 174 | 0.47% |
| **fluid** | 169 | 0.46% |
| **inguinal** | 151 | 0.41% |
| **abdomen** | 142 | 0.39% |
| **bladder** | 131 | 0.36% |
| **bowel** | 114 | 0.31% |
| **wall** | 109 | 0.30% |

Radiologist's Evaluation
Category Subject: Pelvis CT with lesions/tumor
Consistency: High

# Cluster #019

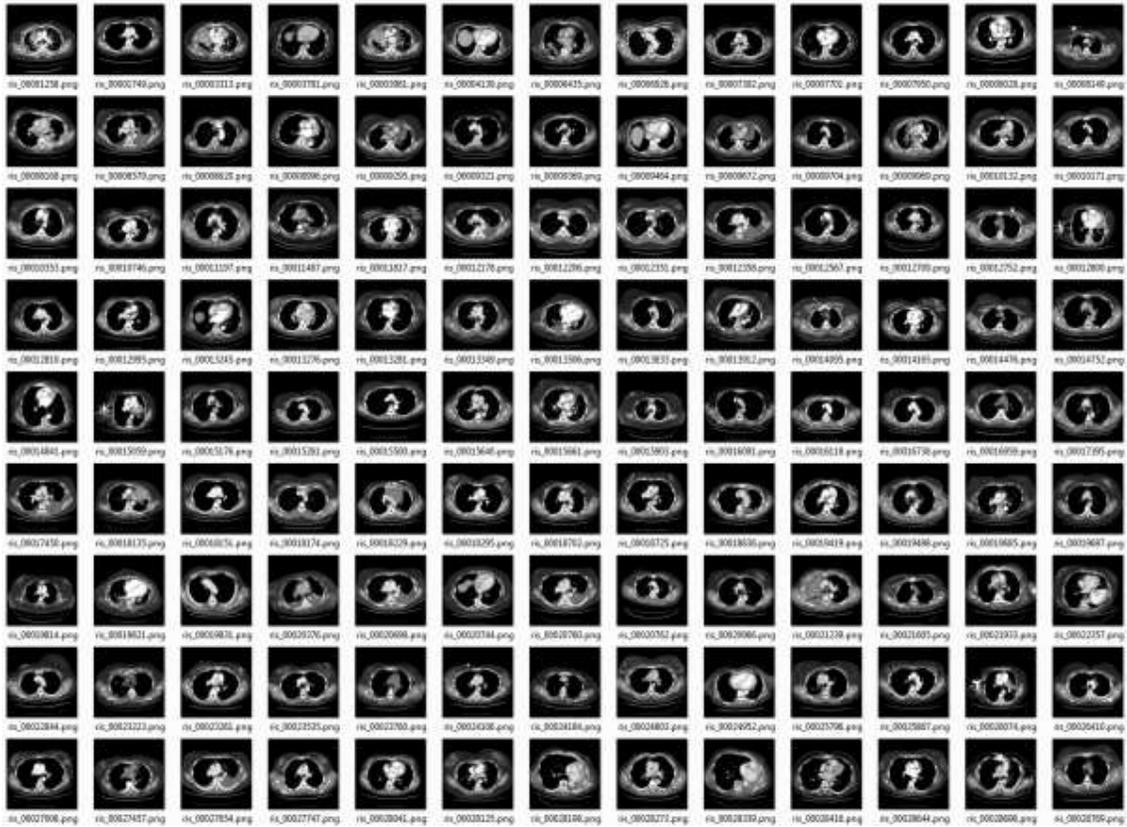

| WORD | Frequency | Relative Frequency |
|---|---|---|
| mass | 324 | 1.05% |
| lung | 322 | 1.05% |
| chest | 318 | 1.03% |
| adenopathy | 307 | 1.00% |
| lobe | 225 | 0.73% |
| pleural | 223 | 0.72% |
| hilum | 148 | 0.48% |
| pulmonary | 142 | 0.46% |
| effusion | 134 | 0.44% |
| axilla | 123 | 0.40% |

Radiologist's Evaluation
Category Subject: Chest CT with pulmonary disease (predominately pleural)
Consistency: High

# Cluster #020

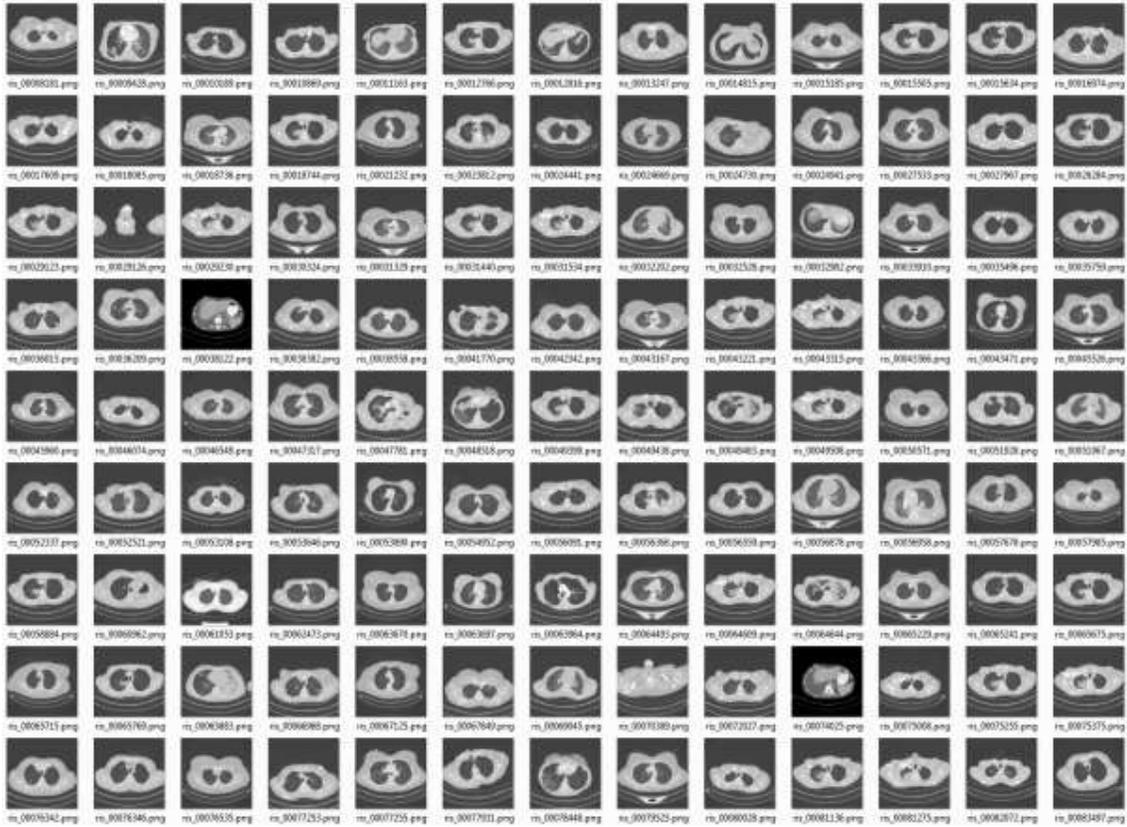

| WORD | Frequency | Relative Frequency |
|---|---|---|
| lung | 883 | 2.13% |
| lobe | 627 | 1.51% |
| nodule | 521 | 1.26% |
| chest | 504 | 1.22% |
| adenopathy | 269 | 0.65% |
| mass | 226 | 0.55% |
| masses | 219 | 0.53% |
| nodules | 205 | 0.50% |
| axilla | 159 | 0.38% |
| scarring | 151 | 0.36% |

Radiologist's Evaluation
Category Subject: Chest CT with tumor
Consistency: High



\end{document}